\documentclass[lettersize,journal]{IEEEtran}
\IEEEpubidadjcol
\usepackage{amsmath,amsfonts}
\usepackage{algorithmic}
\usepackage{algorithm}
\usepackage{array}
\usepackage[caption=false]{subfig}
\usepackage{textcomp}
\usepackage{stfloats}
\usepackage{url}
\usepackage{verbatim}
\usepackage{graphicx}
\usepackage{cite}
\usepackage{multirow}
\usepackage{makecell}
\usepackage{threeparttable}
\usepackage{pifont}
\usepackage{balance}
\usepackage[colorlinks,
	linkcolor=blue,
	anchorcolor=blue,
	citecolor=blue,
	urlcolor=blue]{hyperref}
\usepackage{tikz,xcolor}

\definecolor{lime}{HTML}{A6CE39}
\DeclareRobustCommand{\orcidicon}{
\begin{tikzpicture}
\draw[lime, fill=lime] (0,0)
circle[radius=0.16]
node[white]{{\fontfamily{qag}\selectfont \tiny \.{I}D}};
\end{tikzpicture}
\hspace{-2mm}
}
\foreach \x in {A, ..., Z}{%
\expandafter\xdef\csname orcid\x\endcsname{\noexpand\href{https://orcid.org/\csname orcidauthor\x\endcsname}{\noexpand\orcidicon}}
}

\begin{document}

\title{GraphCFC: A Directed Graph Based Cross-Modal Feature Complementation Approach for Multimodal Conversational Emotion Recognition}

\author{Jiang Li\hspace{-1.5mm}\orcidA{}, Xiaoping Wang\hspace{-1.5mm}\orcidB{},~\IEEEmembership{Senior Member, IEEE}, Guoqing Lv, and Zhigang Zeng\hspace{-1.5mm}\orcidC{},~\IEEEmembership{Fellow, IEEE}\\
        % <-this % stops a space
\thanks{Manuscript received 30 June 2022; revised 12 November 2022 and 26 February 2023; accepted 16 March 2023. This work was supported in part by the National Natural Science Foundation of China under Grant 62236005, 61876209 and 61936004. The Associate Editor coordinating the review of this manuscript and approving it for publication was Dr. Ramanathan Subramanian. \textit{(Corresponding authors: Jiang Li and Xiaoping Wang.)}}% <-this % stops a space
\thanks{The authors are with the School of Artificial Intelligence and Automation and the Key Laboratory of Image Processing and Intelligent Control of Education Ministry of China, Huazhong University of Science and Technology, Wuhan 430074, China (e-mail:lijfrank@hust.edu.cn; wangxiaoping@hust.edu.cn; guoqinglv@hust.edu.cn; zgzeng@hust.edu.cn).}
\thanks{Digital Object Identifier 10.1109/TMM.2023.3260635}}

\markboth{IEEE TRANSACTIONS ON MULTIMEDIA}%
{Li \MakeLowercase{\textit{et al.}}: A Directed Graph Based Cross-Modal Feature Complementation Approach}

\IEEEpubid{1520--9210~\copyright~2023 IEEE. Personal use is permitted, but republication/redistribution requires IEEE permission.}

\maketitle

\begin{abstract}
Emotion Recognition in Conversation (ERC) plays a significant part in Human-Computer Interaction (HCI) systems since it can provide empathetic services. Multimodal ERC can mitigate the drawbacks of uni-modal approaches. Recently, Graph Neural Networks (GNNs) have been widely used in a variety of fields due to their superior performance in relation modeling. In multimodal ERC, GNNs are capable of extracting both long-distance contextual information and inter-modal interactive information. Unfortunately, since existing methods such as MMGCN directly fuse multiple modalities, redundant information may be generated and diverse information may be lost. In this work, we present a directed Graph based Cross-modal Feature Complementation (GraphCFC) module that can efficiently model contextual and interactive information. GraphCFC alleviates the problem of heterogeneity gap in multimodal fusion by utilizing multiple subspace extractors and Pair-wise Cross-modal Complementary (PairCC) strategy. We extract various types of edges from the constructed graph for encoding, thus enabling GNNs to extract crucial contextual and interactive information more accurately when performing message passing. Furthermore, we design a GNN structure called GAT-MLP, which can provide a new unified network framework for multimodal learning. The experimental results on two benchmark datasets show that our GraphCFC outperforms the state-of-the-art (SOTA) approaches.
\end{abstract}


\section{Introduction}
\IEEEPARstart{E}{motions} pervade our personal and professional lives, shape our relationships and social interactions~\cite{van2022social}. Consequently, research on emotion recognition and understanding is crucial. Emotion recognition in conversation (ERC), which aims to automatically determine the emotional state of a speaker during a conversation based on information about human behavior such as text content, facial expressions and audio signals, has received extensive attention and study in recent years~\cite{ghosal2019dialoguegcn,majumder2019dialoguernn,hu2021dialoguecrn,hu2021mmgcn}. Emotion recognition can be applied to many practical scenarios such as medical diagnosis~\cite{huang2019speech}, opinion mining~\cite{chatterjee2019semeval}, fake news detection~\cite{zhang2021mining} and dialogue generation~\cite{huang2018automatic}, to provide high-quality and humanized empathetic services. ERC will play an increasingly vital role as Human-Computer Interaction (HCI) technology advances.

\IEEEpubidadjcol
In a multi-person dialogue scenario, each speaker generates a succession of ups and downs in emotional reactions. The majority of prior techniques have been centered on the study of contextual ERC systems. DialogueGCN~\cite{ghosal2019dialoguegcn} utilizes a relational Graph ATtention network (GAT) to capture long-distance contextual dependencies in conversations, and leverages self- and inter-dependency of the speakers to improve context understanding for ERC. Shen et al.~\cite{DBLP:conf/acl/ShenWYQ20} model the conversation as a directed acyclic graph in an attempt to combine the respective strengths of Graph Neural Networks (GNNs) and Recurrent Neural Networks (RNNs). DialogueCRN~\cite{hu2021dialoguecrn} designs multi-turn reasoning modules to extract and integrate the emotional clues in order to fully understand the conversational context from a cognitive perspective. Nonetheless, these approaches only take into account information from a single modality. The ERC system will fail if the uni-modality signals do not carry a significant emotional signature. Furthermore, the notorious emotional-shift issue plagues uni-modal emotion recognition systems~\cite{majumder2019dialoguernn,DBLP:conf/acl/ShenWYQ20}.

\begin{figure}[htbp]
\centering
\includegraphics[width=3.4in]{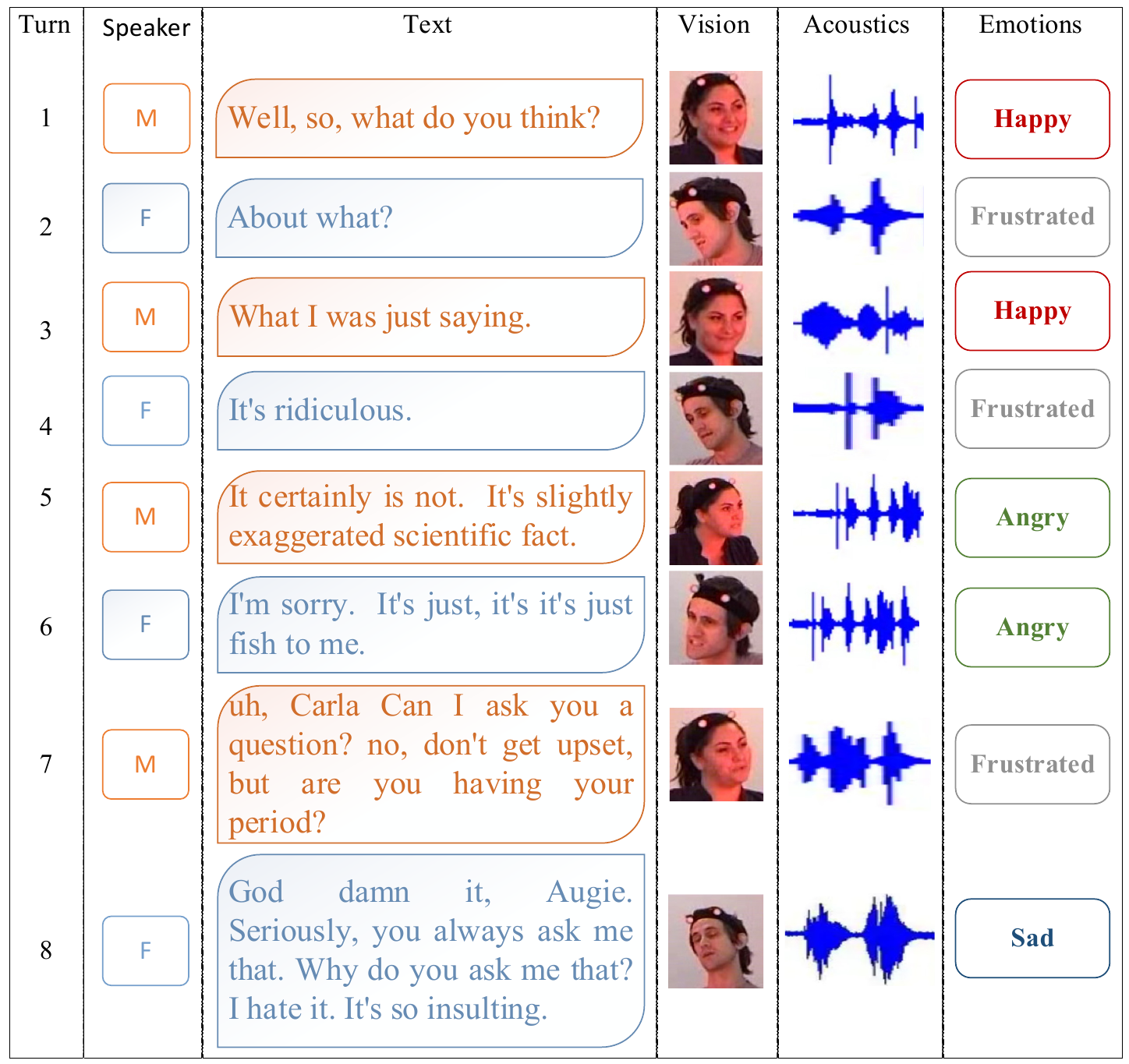}
\caption{An instance of a multimodal dialogue system. The utterances contain three modalities, i.e., textual, acoustic and visual modalities.}
\label{fig:multi_dialogue}
\end{figure}

In real scenarios, people can instinctively obtain complex emotional cues from multiple information sources including the facial expressions, personality and tone of speaker, as well as the conversation history to infer the real emotions of others. Multimodal ERC follows this idea and attempts to combine simultaneously information from multiple modalities such as textual, acoustic and visual modalities to comprehensively identify emotions in conversations. Fig.~\ref{fig:multi_dialogue} shows an instance of a multimodal conversation system. The ERC system takes each modality as input and then performs emotional prediction. bc-LSTM~\cite{poria2017context} employs textual, visual and acoustic modalities for multimodal emotion recognition, feeding each modality separately into a bidirectional Long Short-Term Memory (LSTM) network to acquire contextual information. CMN~\cite{hazarika2018conversational} uses Gated Recurrent Unit (GRU) and multimodal features for contextual modeling, as well as applies an attention mechanism to pick the most valuable historical utterances. ICON~\cite{hazarika2018icon} models the contextual knowledge of self- and inter-speaker impacts via a GRU-based multi-hop memory network, while capturing essential emotional cues applying an attention module. DialogueRNN~\cite{majumder2019dialoguernn} detects current sentiment by tracking the contextual information of the utterance and considering the characteristics of the speaker. These approaches, nevertheless, directly concatenate multimodal information without incorporating the interaction between modalities. In addition, recurrence-based approaches tend to use recent utterances for modeling, which makes these models difficult to gather long-distant information.

Recently, GNNs have attracted wide attention in a variety of fields because they can model relationships. MMGCN~\cite{hu2021mmgcn} achieves outstanding performance in multimodal conversational emotion recognition by employing GNNs to capture long-distance contextual information and inter-modal interactive information. However, MMGCN connects the current node directly to all other nodes in the dialogue, perhaps resulting in redundant information. Besides that MMGCN simply divides all edges into two types (i.e., inter-modal edges and intra-modal edges) and leverages the angular similarity to represent edge weight, which can cause the inability of the GNN to accurately select important information when aggregating neighbor information.

There is a heterogeneity gap~\cite{guo2019dmrl,hazarika2020misa} between distinct modalities in multimodal fusion, which makes it challenging to effectively fuse multimodal features. MMGCN directly puts the utterance of three modalities into the graph as the same type of nodes, and then performs multimodal feature fusion by GNN. This approach not only adds redundant information due to inconsistent data distribution among modalities, but also may risk losing diverse information in the conversational graph. Therefore, we propose a novel graph-based multimodal feature fusion method to alleviate the aforementioned limitations. In the Graph based Cross-modal Feature Complementation (GraphCFC) module, unlike MMGCN treating all utterances as neighbor nodes, we model the conversation as a multimodal directed heterogeneous graph with variable contextual information and extract more than two types of edges from the graph based on the perspective of modality type and speaker identity; then, we utilize multiple subspace extractors to simultaneously preserve the consistency and diversity of multimodal features; finally, we employ the Pair-wise Cross-modal Complementation (PairCC) strategy to gradually achieve feature complementation and fusion. In addition, we propose a novel GNN layer, GAT-MLP, to provide a unified network model for multimodal feature fusion, which can also effectively minimize the over-smoothing problem~\cite{li2018deeper} of GNNs. Our main contributions in this paper are as follows:
\begin{itemize}
\item[\ding{172}] We propose a directed Graph based Cross-modal Feature Complementation (GraphCFC) module. GraphCFC can not only effectively alleviate the heterogeneity gap issue of multimodal fusion, but also sufficiently extract the diverse information from multimodal dialogue graphs.
\item[\ding{173}] A new GNN layer named GAT-MLP is designed, which not only alleviates the over-smoothing problem of GNNs, but also provides a new network framework for multimodal learning.
\item[\ding{174}] The conversations are represented as a multimodal directed graph with variable contextual utterances and extract distinct types of edges from this graph for encoding, so that GAT-MLP can accurately select the critical contextual and interactive information.
\item[\ding{175}] Extensive comparative experiments and ablation studies are conducted on two benchmark datasets. The experimental results reveal that our proposed GraphCFC is capable of productive complementation and fusion of multimodal features, attaining optimal performance in comparison to previous SOTA approaches.
\end{itemize}

The remainder of this paper is organized as follows. The related works of this paper is briefly mentioned in Section \ref{sec:related_work}. Section \ref{sec:proposed_methods} depicts the proposed graph-based multimodal ERC method. Section \ref{sec:experiment} presents the experimental setup of this work, and the experimental results are analyzed in detail in Section \ref{sec:results_analysis}. Section \ref{sec:conclusion} summarizes and prospects to this work.

\section{Related Work}\label{sec:related_work}
\subsection{Emotion Recognition in Conversation}
Emotion Recognition in Conversation (ERC), which aims to predict the emotion label of each utterance, plays a crucial role in affective dialogue due to facilitating the understanding of the user's emotions and responding with empathy. This task has been recently attached much importance by numerous NLP researchers for its potential applications in extensive areas such as opinion mining in social media~\cite{chatterjee2019semeval}, empathy building in dialogue systems~\cite{majumder2020mime} and detection of fake news~\cite{zhang2021mining}. The emotion of a query utterance is easily influenced by many factors, such as the identity of speaker and the context of conversation. Indeed, the way to model the conversational context is the core of this task~\cite{poria2018meld}.

Massive methods have been taken to model the conversation context on the textual modality, which can be divided into two categories: graph-based methods and recurrence-based methods. Besides, models based on multimodal inputs have been proposed, which improve the performance of ERC tasks by leveraging multimodal dependencies and complementarities.

\textit{Graph-based methods:} DialogGCN~\cite{ghosal2019dialoguegcn} constructs a dialogue graph where each utterance is related with the surrounding utterances. Ishiwatari et al.~\cite{ishiwatari2020relation} improves DialogGCN by taking positional encoding into account. ConGCN~\cite{zhang2019modeling} constructs a large heterogeneous graph by treating the speakers and utterances as nodes. KET~\cite{zhong2019knowledge} leverages a context-aware affective graph attention mechanism to dynamically capture external commonsense knowledge. DAG-ERC~\cite{DBLP:conf/acl/ShenWYQ20} combines the advantages of both graph neural networks and recurrent neural networks, and performs excellently without the aid of external knowledge.

\textit{Recurrence-based methods:} ICON~\cite{hazarika2018icon} and CMN~\cite{hazarika2018conversational} both utilize Gated Recurrent Unit (GRU) and memory networks. HiGRU~\cite{jiao2019higru} is made up of two GRUs, one is an utterance encoder and the other is a conversation encoder. DialogRNN~\cite{majumder2019dialoguernn} is a sequence-based method, where several RNNs model the dialogue dynamically. COSMIC~\cite{ghosal2020cosmic} constructs a network that is closely similar to DialogRNN and performs better by adding external commonsense knowledge. DialogueCRN~\cite{hu2021dialoguecrn} utilizes bidirectional LSTM to build ERC model from a cognitive perspective.

\textit{Multimodal-based methods:} CMN~\cite{hazarika2018conversational} leverages multimodal information by concatenating the features from three modalities but fails to consider the interaction between modalities. bc-LSTM~\cite{poria2017context} adopts an utterance-level LSTM to capture multimodal information. MFN~\cite{sahay2018multimodal} conducts multi-views information fusion and aligns the features of different modalities, but it is unable to model speaker information. MMGCN~\cite{hu2021mmgcn} utilizes an undirected graph to explore a more effective way of multimodal fusion, which outperforms significantly other approaches under the multimodal dialogue setting. There have been a range of works~\cite{tsai2018learning,tsai2019multimodal,hazarika2020misa,sun2020learning} associated with multimodal learning in sentiment analysis. These efforts, however, do not highlight the social interaction and contextual cues between the speaker and listener in a conversation, thus they do not fall under the purview of ERC. In addition, most sentiment analysis tasks only need to distinguish positive, negative, and neutral opinions. Thus it is difficult to divide emotion into numerous categories like \textit{Happy}, \textit{Excited}, \textit{Sad}, \textit{Angry}, \textit{Frustrated} as in the case of ERC tasks.

\subsection{Multimodal Fusion}
Multimodal fusion is one of the most important parts in machine learning, which can integrate information from multiple modalities to predict a result~\cite{baltruvsaitis2018multimodal}. It can be intuitively assumed that multimodal fusion can provide three benefits: providing more robust prediction results, capturing complementary information, and working without certain modalities~\cite{d2015review}. In recent years, multimodal fusion in sentiment analysis~\cite{tsai2018learning,tsai2019multimodal,sun2020learning} has been researched widely and lots of methods such as multiple kernel learning and various neural networks have been explored to cope with it. However, multimodal sentiment analysis rarely involves multi-person conversational information and focuses mainly on utterance-level prediction. Multimodal ERC is the study of conversations with two or more participants and is a conversation-level emotion prediction. For instances, an individual's emotion is not only derived from self-expression, but is also influenced by the expressions of others. Furthermore, Guo et al.~\cite{guo2019dmrl} have noted that multimodal fusion faces several challenges, one of which is the heterogeneity gap~\cite{hazarika2020misa} between modalities. For this reason, our proposed GraphCFC concentrates on alleviating the heterogeneity gap dilemma of conversational emotion recognition.

\subsection{Graph Neural Networks}
In recent years, an increasing number of non-Euclidean data have been represented as graphs. The complexity of graph data has posed significant challenges to existing neural network models. Graph Neural Networks (GNNs) have attracted much attention for their ability to effectively deal with non-Euclidean data. GNNs have been applied in a wide range of applications, including recommendation system, computer vision, natural language processing, biomedicine and traffic forecasting. Graph convolutional networks, a type of GNNs, can be divided into two main streams: the spectral-based and the spatial-based approaches. Spectral-based approaches implement graph convolution by defining filters in a manner similar to graph signal processing. Spatial-based approaches define graph convolution by information propagation, and they have recently gained rapid momentum due to their attractive efficiency, flexibility, and generality. Graph-SAGE~\cite{hamilton2017inductive}, GAT~\cite{velickovic2017graph}, and FastGCN~\cite{chen2018fastgcn} are widely-used GNN techniques.

\section{Proposed Methods}\label{sec:proposed_methods}
Given the challenges of multimodal emotion recognition mentioned above, we introduce a novel graph-based multimodal feature fusion approach for ERC in this section. The section consists of four parts, including general overview, uni-modal encoder, Graph based Cross-modal Feature Complementation (GraphCFC) module, and multimodal emotion classifier.

\subsection{General Overview}
\subsubsection{Problem Definition}
In an ERC scenario, a dialogue is defined as a sequence of $n$ utterances $[(u_1, s_{u_1}), (u_2, s_{u_2}), ...,\\ (u_n, s_{u_n})]$. Where $n$ denotes the number of utterances, $u_i$ is the $i$-th utterance in this dialogue sequence, and $s_{u_i}$ indicates the speaker who utters utterance $u_i$. Each utterance $u_i$ consists of $m_i$ tokens, i.e., $u_i$ can be expressed as $[t_{i1}, t_{i2}, ..., t_{im_i}]$. The number of speaker $s$ in a dialogue system should be greater than or equal to 2; if $s_{u_i} = s_{u_j}(i \neq j)$, then utterance $u_i$ and $u_j$ are uttered by the same participant in the dialogue. Each utterance $u$ involves three modalities, including textual, acoustic and visual modalities, so utterance $u_i$ can also be denoted as $[u_i^t, u_i^a, u_i^v]$, where $t, a, v$ denote textual, acoustic and visual modalities, respectively. Given the defined emotion labels $Y = [y_1, y_2, ..., y_l]$ ($y_i$ is generally represented by one-hot encoding), the objective of the multimodal ERC task is to predict the emotion state label $y_i$ for each utterance $u_i$ based on the available inter-modal interaction and intra-modal context. The quantity of emotion labels in various datasets varies, e.g., 6 for IEMOCAP and 7 for MELD. We also experimented with coarsened emotion labels, which consisted of \textit{Positive}, \textit{Negative} and \textit{Neutral}. For instance, in the IEMOCAP dataset, \textit{Happy} and \textit{Excited} are categorized as \textit{Positive}; \textit{Sad}, \textit{Angry} and \textit{Frustrated} are categorized as \textit{Negative}; and \textit{Neutral} remained unchanged.

\subsubsection{Overall Architecture}
\begin{figure*}[htbp]
\centering
\includegraphics[width=7.1in]{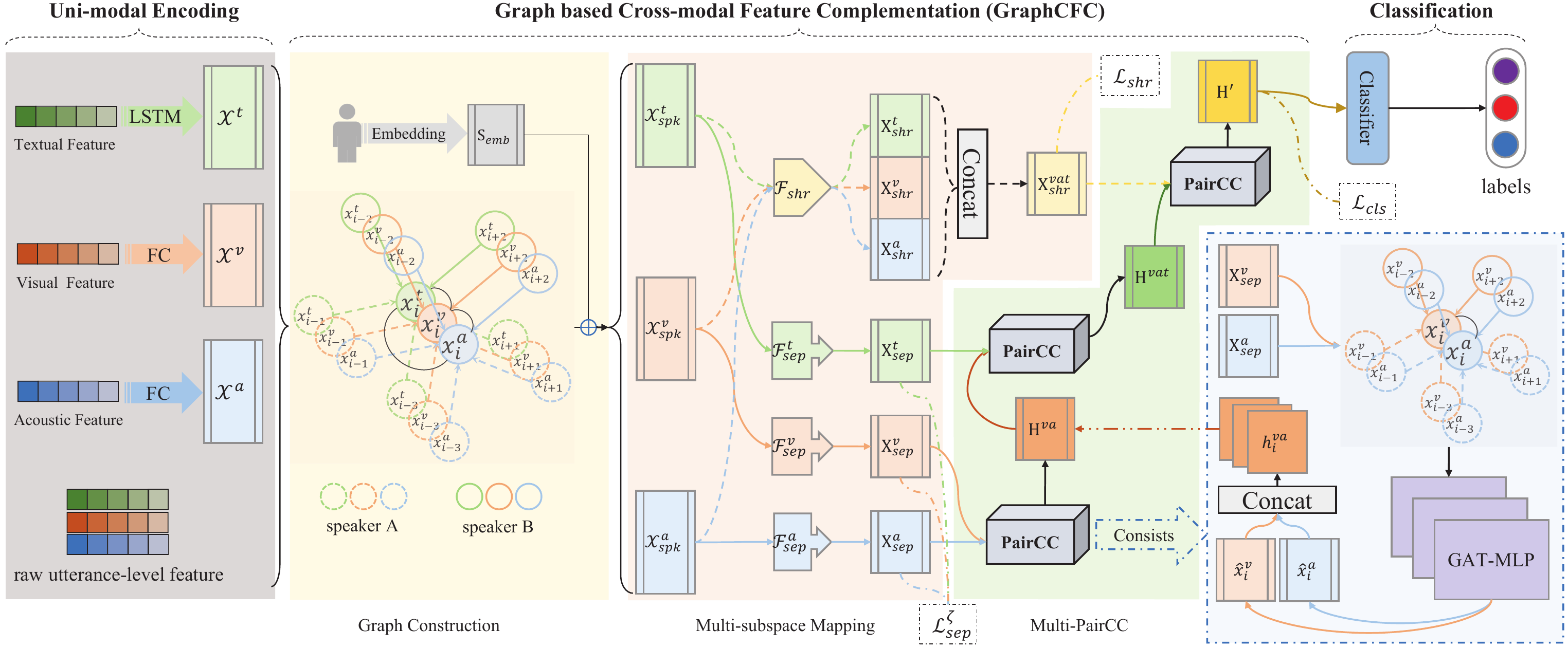}%
\caption{The illustration of graph-based multimodal ERC, which includes uni-modal encoding, graph based cross-modal feature complementation and multimodal emotion classification.}
\label{fig:overall}
\end{figure*}

Fig.~\ref{fig:overall} shows the overall architecture of graph-based multimodal ERC in this paper, which mainly consists of uni-modal encoding, Graph based Cross-modal Feature Complementation (GraphCFC) and multimodal emotion classification. Firstly, we encode the uni-modal features by means of three uni-modal encoders. Next, a cross-modal feature complementation module based on GNN is employed for collecting long-distance intra-modal contextual information and inter-modal interactive information. Finally, we utilize multiple loss functions to build multitask learning model for multimodal emotion classification.

\subsection{Uni-Modal Encoder}
To capture the context-aware feature information of textual modality, referring to MMGCN~\cite{hu2021mmgcn}, we leverage a bidirectional Long Short Term Memory (BiLSTM) network. The feature pre-extraction of textual modality can be formulated as:
\begin{equation}
\label{eq:bilstm}
x_i^t, x_{h,i}^t = \overleftrightarrow{\mathrm{LSTM}}(u_i^t; \mathrm{\Theta}_{ls}^t),
\end{equation}
where $x_i^t$ and $x_{h,i}^t$ are the output and hidden vector of pre-extractor, respectively; $\overleftrightarrow{\mathrm{LSTM}}$ and $\mathrm{\Theta}_{ls}^t$ denote the BiLSTM network and trainable parameter, respectively.

For acoustic and visual modalities, again as with MMGCN, we use a fully connected network for uni-modal feature pre-extraction as follows:
\begin{equation}
\label{eq:fullcon}
x_i^\tau = \mathrm{FC}(u_i^\tau; \mathrm{\Theta}_{fc}^\tau), \tau \in \{a, v\},
\end{equation}
where $x_i^\tau$ is the output vector of pre-extractor; $\mathrm{FC}$ and $\mathrm{\Theta}_{fc}^\tau$ are the fully connected network and trainable parameter, respectively; $a$ and $v$ denote acoustic and visual modalities, respectively.

\subsection{Graph Based Cross-Modal Feature Complementation Module}
We propose a Graph based Cross-modal Feature Complementation (GraphCFC) module for efficient multimodal feature fusion. The module includes two types of information complementation, i.e., intra-modal contextual information and inter-modal interactive information. The four primary reasons for which GraphCFC is proposed are as follows. First, to simultaneously preserve the consistency and diversity information of multimodal features. Second, to select crucial intra-modal contextual information and inter-modal interaction information as accurately as possible. Third, to alleviate the heterogeneity gap problem of multimodal ERC. Last, to propose a network model that can be applied to visual, acoustic, and textual modalities simultaneously.

The GraphCFC Module is divided into five main parts. First, we describe how to construct the graph; second, we introduce multiple subspace mappings which are leveraged to simultaneously ensure the consistency and diversity of multimodal features; third, we present a new graph neural network structure named GAT-MLP; fourth, we introduce GAT-MLP based Pair-wise Cross-modal Complementation (PairCC) for alleviating the heterogeneity gap issue of multimodal ERC; finally, we detail the GAT structure of GAT-MLP used in this work.

\subsubsection{Graph Construction}
In uni-modal ERC, a dialogue with $n$ utterances is represented as a directed graph $\mathcal{G} = (\mathcal{V}, \mathcal{E})$; where $\mathcal{V}$ is the node set, which denotes the set of utterances, i.e., $\mathcal{V} = \{u_1, u_2, ..., u_n \}$; $\mathcal{E}$ is the set of relational dependencies between nodes; and if an edge exists for two nodes, then $e_{ij} \in \mathcal{E}$ has two key properties: edge weight and edge type. Assuming the existence of two modalities $\mathrm{P}, \mathrm{Q}$, we construct the dialogue graph as follows.

\textit{Nodes:} In a graph, each utterance $u_i$ ($i=1, 2,..., n$) is considered as node $u_i^\mathrm{P}$ and node $u_i^\mathrm{Q}$, represented as vector $x_i^\mathrm{P}$ and vector $x_i^\mathrm{Q}$. If there are $n$ utterances, then $\mathcal{V}$ can be denoted as $\mathcal{V}=\{u_1^\mathrm{P},u_1^\mathrm{Q},u_2^\mathrm{P},u_ 2^\mathrm{Q},...,u_n^\mathrm{P},u_n^\mathrm{Q}\}$, $|\mathcal{V}|=2 \times n$. In $M$ modalities, $|\mathcal{V}|=M \times n$, $M$ is the number of modalities and $n$ is the number of utterances.

\textit{Edges:} In a graph, an edge is defined as a connection between nodes. In the dialogue graph of multiple modalities, we define edges in two perspectives: the contextual connection of intra-modal utterance, and the interactive connection of inter-modal utterance. Particularly, we term these two types of edges as intra-edge ($\mathcal{E}_{intra}$) and inter-edge ($\mathcal{E}_{inter}$), respectively. The intra-edge is utilized for capturing intra-modal contextual information, whereas the inter-edge is utilized for capturing cross-modal interactive information.

The intra-edge is defined as follows. Assuming the existence of modality $\mathrm{P}$,  we connect the current utterance node $u_i^\mathrm{P}$ with the previous/past $j$ utterance nodes $u_{i-j}^\mathrm{P},u_{i-j+1}^\mathrm{P},...,u_{i-1}^\mathrm{P}$. Similarly, we connect $u_i^\mathrm{P}$ with the next/future $k$ utterance nodes $u_{i+1}^\mathrm{P},u_{i+2}^\mathrm{P},...,u_{i+k}^\mathrm{P}$. Therefore, we can formalize $\mathcal{E}_{intra}$ as follows: 
\begin{equation}
\label{eq:intra-edge}
\mathcal{E}_{intra} =
\begin{cases}
\{(u_t^\mathrm{P}, u_i^\mathrm{P}) | i-j < t < i-1 \}\\
\{(u_i^\mathrm{P}, u_t^\mathrm{P}) | i+1 < t < i+k \} 
\end{cases},
\end{equation}
where $i$, $j$, $k$ are constants, $t$ is a variable; and $i$, $j$, $k$ are less than $n$; $i$, $j$, $k$, $t$ all belong to $\mathbb{N_+}$.

The inter-edge is defined as follows. In a dialogue, we connect the utterance node $u_i^\mathrm{P}$ of modality $\mathrm{P}$ to the corresponding utterance node $u_i^\mathrm{Q}$ of modality $\mathrm{Q}$. Thus, we can formulate $\mathcal{E}_{inter}$ as follows:
\begin{equation}
\label{eq:inter-edge}
\mathcal{E}_{inter} = \{(u_i^\mathrm{P}, u_i^\mathrm{Q}), (u_i^\mathrm{Q}, u_i^\mathrm{P})\},
\end{equation}
where $i < n$ and $i \in \mathbb{N_+}$.

\textit{Edge types:} Based on the definition of edges above, we may divide all edges into two types: intra-edge type and inter-edge type, labeled as $\mathrm{ET}_{intra}$ and $\mathrm{ET}_{inter}$, respectively. It is commonly known that if two utterance nodes in a dialogue has edge, it may or may not be from the same speaker. Therefore, $\mathrm{ET}_{intra}$ can be subdivided according to the perspective of speaker. Specifically, suppose that there are 3 speakers ($s_1$, $s_2$, $s_3$) in a dialogue, then the set of existing edge types when only a single modality is considered can be written as:
\begin{equation}
\label{eq:et-intra}
\begin{split}
\mathrm{ET}_{intra} = \{\mathrm{et}(s_1,s_1),\mathrm{et}(s_1,s_2),\mathrm{et}(s_1,s_3),\\ 
\mathrm{et}(s_2,s_2),\mathrm{et}(s_2,s_3),\mathrm{et}(s_3,s_3)\}.
\end{split}
\end{equation}
It can be easily concluded that if there are $D$ speakers in a dialogue, then there are $D \times (D + 1)/2$ kinds of edges/relations. If there are $M$ modalities, then there are $M \times (D^2 + D)/2$ elements in $\mathrm{ET}_{intra}$. 

$\mathrm{ET}_{inter}$ denotes the set of inter-modal edge types. Suppose that there are 3 modalities ($mod_1$, $mod_2$, $mod_3$) of the same utterance, then $\mathrm{ET}_{inter}$ can be formalized as:
\begin{equation}
\label{eq:et-inter}
\begin{split}
\mathrm{ET}_{inter} = \{\mathrm{et}(mod_1,mod_2),\mathrm{et}(mod_1,mod_3),\\
\mathrm{et}(mod_2,mod_3)\}.
\end{split}
\end{equation}
Therefore, if there are $M$ modalities in the same utterance, then existing $M \times (M-1)/2$ kinds of edges/relations. In this work, we consider three modalities of an utterance, so that there are a total of three elements in $\mathrm{ET}_{inter}$.

\textit{Edge weights:} Edge weights are utilized to identify the relevance of distinct neighboring nodes when GNNs aggregates information. We employ a learnable attention module with edge features, which is detailed in Section \ref{subsubsec:singlegat}.

\subsubsection{Multi-Subspace Extractor}
Inspired by MMGCN~\cite{hu2021mmgcn}, we consider the speaker information is of importance. The embedding of multi-speaker $S_{emb}$ can be formalized as:
\begin{equation}
\label{eq:emb-spk}
S_{emb} = \mathrm{Embedding}(\mathrm{S}, D),
\end{equation}
where $\mathrm{S}$ denotes the set of speakers, $D$ is the number of speakers. To encode the speaker identity information, we add the speaker embedding to the features of utterance nodes:
\begin{equation}
\label{eq:add-spk}
\mathcal{X}_{spk}^\zeta = \mu S_{emb} + \mathcal{X}^\zeta,
\end{equation}
where $\mathcal{X}^\zeta$ ($\zeta \in \{t, a, v\}$) is the feature matrix from uni-modal encoder, and $x_i^\zeta \in \mathcal{X}^\zeta$; $\mathcal{X}_{spk}^\zeta$ denotes the feature matrix adding the speaker embedding; $\mu \in [0,1]$ is the ratio of the speaker embedding.

Currently, one of the quandaries we confront in multimodal fusion is the existence of heterogeneity gap~\cite{guo2019dmrl}. In other words, the distribution of data is inconsistent across modalities. Therefore, before performing cross-modal feature complementation, we map the features of each modality into a shared subspace to maintain the consistency of feature representation across modalities. But yet the more similar the feature representations of multiple modalities are, the less complementary the feature between modalities are. To put it another way, we want to preserve the diversity of feature representations of multiple modalities so that the features of one modality can complement those of others. In view of this, we map the features of each modality into separate subspaces for capturing the diversity of feature representations across modalities. We argue that capturing the diversity and consistency information of multiple modalities simultaneously facilitates the complementation and fusion between modalities.

In order to capture the consistency of multimodal information in the shared subspace, we use three mapping functions $\mathcal{F}_{shr}$ with the same trainable parameter $\mathrm{\Theta}_{shr}$. For the separate subspace, we use three mapping functions $\mathcal{F}_{sep}$ with different trainable parameters $\mathrm{\Theta}_{sep}^\zeta$ to capture the variety of multimodal information. The two kinds of mapping methods are shown in Fig.~\ref{fig:overall}, and are formulated as follows:
\begin{equation}
\label{eq:shr-sep}
\begin{split}
&\mathrm{X}_{shr}^\zeta = \mathcal{F}_{shr}(\mathcal{E}_{intra}^\zeta, \mathcal{X}_{spk}^\zeta; \mathrm{\Theta}_{shr}), \\
&\mathrm{X}_{shr}^{vat} = \mathrm{Lin}([\mathrm{X}_{shr}^v \, \Vert \, \mathrm{X}_{shr}^a \, \Vert \, \mathrm{X}_{shr}^t]; \mathrm{\Theta}'_{shr}),\\
&\mathrm{X}_{sep}^\zeta = \mathcal{F}_{sep}(\mathcal{E}_{intra}^\zeta, \mathcal{X}_{spk}^\zeta; \mathrm{\Theta}_{sep}^\zeta),
\end{split}
\end{equation}
where $\Vert$ denotes concatenation operation; $\mathrm{X}_{shr}^\zeta$ ($\mathrm{X}_{sep}^\zeta$) denotes the consistency (diversity) feature matrices, and $\zeta \in \{t, a, v\}$; $\mathrm{\Theta}_{shr}$, $\mathrm{\Theta}'_{shr}$ and $\mathrm{\Theta}_{sep}^\zeta$ are the trainable parameters. Note that the mapping function $\mathcal{F}$ can be a fully connected layer, a graph neural network layer, etc. In this paper, we define the mapping function $\mathcal{F}$ as follows:
\begin{equation}
\label{eq:mapping-function}
\begin{split}
&\mathcal{F}(\mathcal{X}; \mathrm{\Theta}) = \\
&\mathrm{Norm}(\mathrm{Drop}(\mathrm{Lin}(\mathrm{Drop}(\mathrm{\sigma}(\mathrm{Lin}(\mathcal{X}; \Theta_{0}))); \Theta_{1}))),            
\end{split}
\end{equation}
the mapping function $\mathcal{F}$ is actually two fully connected layers; where $\mathcal{X}$ is the input of $\mathcal{F}$; $\mathrm{Lin}$, $\mathrm{\sigma}$, $\mathrm{Drop}$ and $\mathrm{Norm}$ denote the linear, non-linear activation, dropout and normalization functions, respectively; $\Theta$ denotes the learnable parameter. 

Despite the fact that the shared mapping function $\mathcal{F}_{shr}$ and the separate mapping function $\mathcal{F}_{sep}$ are utilized to extract consistency and diversity features, they should have the equivalent learning goal, i.e., the features of the same utterance mapped by different functions should correspond to the same emotion label. Therefore, we utilize four subspace loss functions to limit the features extracted by muti-subspace extractor such that they do not deviate from the ultimate goal task. The shared subspace loss function is computed as:
\begin{equation}
\label{eq:shr-loss0}
\begin{split}
&v'_i = \mathrm{ReLU}(\mathrm{W}'_0 \mathrm{x}_{shr,i}^{vat}+\mathrm{b}'_0),\\
&p'_i = \mathrm{Softmax}(\mathrm{W}'_1 v_i^s+\mathrm{b}'_1),
\end{split}
\end{equation}
\begin{equation}
\label{eq:shr-loss}
\mathcal{L}_{shr} = - \frac {1}{\sum_{k=0}^{N-1} n(k)} \sum_{i=0}^{N-1}\sum_{j=0}^{n(i)-1} y_{ij} \log p'_{ij}
+ \lambda \lvert \mathrm{\Theta}'_{re} \rvert,
\end{equation}
where $\mathrm{x}_{shr,i}^{vat} \in \mathrm{X}_{shr}^{vat}$; $N$ is the number of dialogues, $n(i)$ is the number of utterances in dialogue $i$; $y_{ij}$ denotes the ground truth label of the $j$-th utterance in the $i$-th dialogue, $p'_{ij}$ denotes the probability distribution of predicted emotion label of the $j$-th utterance in the $i$-th dialogue; $\lambda$ is the L2-regularizer weight, and $\mathrm{W}'_0$, $\mathrm{W}'_1$, $\mathrm{b}'_0$, $\mathrm{b}'_1$, $\mathrm{\Theta}'_{re}$ are the trainable parameters. Similarly, the separate subspace loss function is computed as:
\begin{equation}
\label{eq:sep-loss0}
\begin{split}
&v_i^{\zeta} = \mathrm{ReLU}(\mathrm{W}_0^{\zeta} \mathrm{x}_{sep,i}^{\zeta}+\mathrm{b}_0^{\zeta}),\\
&p_i^{\zeta} = \mathrm{Softmax}(\mathrm{W}_1^{\zeta} v_i^{\zeta}+\mathrm{b}_1^{\zeta}),
\end{split}
\end{equation}
\begin{equation}
\label{eq:sep-loss}
\mathcal{L}_{sep}^{\zeta} = - \frac {1}{\sum_{k=0}^{N-1} n(k)} \sum_{i=0}^{N-1}\sum_{j=0}^{n(i)-1} y_{ij} \log p_{ij}^{\zeta} 
+ \lambda \lvert \mathrm{\Theta}_{re}^{\zeta}\rvert, 
\end{equation}
where $\mathrm{x}_{sep,i}^{\zeta} \in \mathrm{X}_{sep}^{\zeta}$, $\zeta \in \{t, a, v\}$ is the type of modalities, i.e., textual, acoustic and visual modalities; $\mathrm{W}_0^{\zeta}$, $\mathrm{W}_1^{\zeta}$, $\mathrm{b}_0^{\zeta}$, $\mathrm{b}_0^{\zeta}$, $\mathrm{\Theta}_{re}^{\zeta}$ are the learnable parameters.

\begin{figure*}[htbp]
\centering
\includegraphics[width=7.0in]{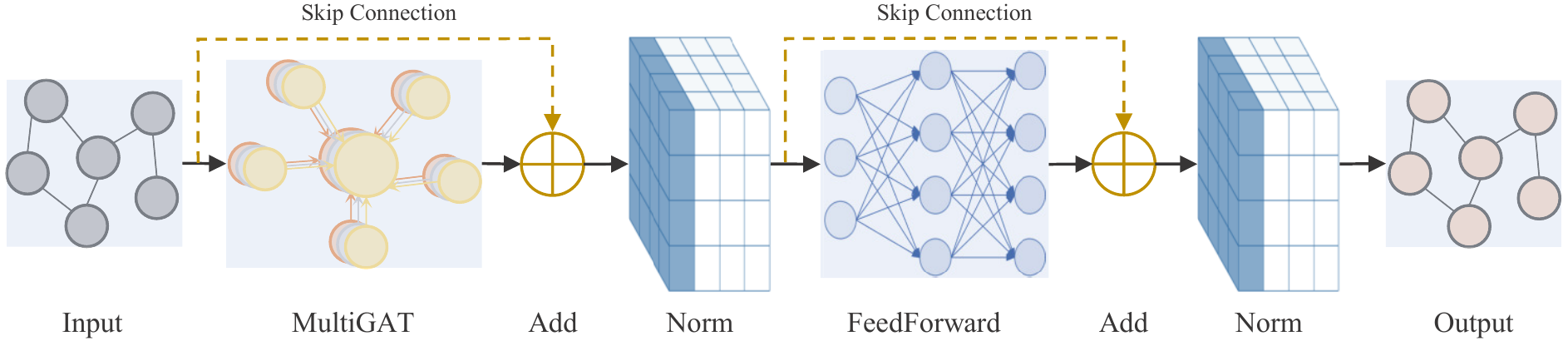}%
\caption{The structure of the designed GAT-MLP, where the $\mathrm{Norm}$ operation can be placed before $\mathrm{MultiGAT}$ and $\mathrm{FeedForward}$. $\mathrm{MultiGAT}$ denotes the multi-head graph attention network; the layer normalization function is used for the $\mathrm{Norm}$ in our work.}
\label{fig:gat_mlp}
\end{figure*}

\subsubsection{GAT-MLP Layer}
It is well known that the dilemma of over-smoothing exists in GNNs. Experimental studies have shown that the performance of the model begins to deteriorate dramatically as the number of layers in the GNN reaches a specific threshold. This is due to the impact of graph convolutions in that it inherently makes representations of adjacent nodes closer to each other~\cite{li2018deeper}. Thus, after multiple graph convolutions, the node features within the same connected component of the network tend to be similar and the model degenerates.

Inspired by the ResNet~\cite{he2016deep} model, ResGCN~\cite{li2021deepgcns} was proposed to address the over-smoothing of GNNs. ResGCN and most of the extant others, however, concatenate the output of each layer, which potentially limit the expressiveness of the model. Transformer has seen success in the fields of computer vision~\cite{dosovitskiy2020image, liu2021swin}, natural language processing~\cite{vaswani2017attention}, and speech recognition~\cite{zhang2020transformer} in recent years, and its network structure is regarded as excellent. As depicted in Fig.~\ref{fig:gat_mlp}, we design a new GNN layer called GAT-MLP based on the ideas of ResNet and Transformer. The GAT-MLP layer can be formulated as:
\begin{equation}
\label{gat-mlp}
\begin{split}
&\mathrm{X}_\mathrm{gat} = \mathrm{Norm}(\mathrm{MultiGAT}(\mathcal{E}, \mathrm{X}_{in}; \mathrm{\Theta}_\mathrm{gat}) + \mathrm{X}_{in}),\\
&\mathrm{X}_{out} = \mathrm{Norm}(\mathrm{FeedForward}(\mathrm{X}_\mathrm{gat}; \mathrm{\Theta}_\mathrm{fed}) + \mathrm{X}_\mathrm{gat}),
\end{split}
\end{equation}
where $\mathrm{X}_{in}$ ($\mathrm{X}_{out}$) denotes the input (output) matrix of node features; $\mathcal{E}$ denotes the set of edge; $\mathrm{MultiGAT}$ is the multi-head graph attention network; $\mathrm{\Theta}_\mathrm{gat}$ and $\mathrm{\Theta}_\mathrm{fed}$ are the trainable parameters; $\mathrm{FeedForward}$ and $\mathrm{Norm}$ are the feedforward and normalization functions, respectively. The layer normalization function is used as $\mathrm{Norm}$ in this work. The feedforward function is computed as follows:
\begin{equation}
\label{feedforward}
\begin{split}
&\mathrm{FeedForward}(\mathrm{X}_\mathrm{gat}, \mathrm{\Theta}_\mathrm{fed}) = \\
&\mathrm{Drop}(\mathrm{Lin}(\mathrm{Drop}(\mathrm{\sigma}(\mathrm{Lin}(\mathrm{X}_\mathrm{gat}; \Theta_0))); \Theta_1)),
\end{split}
\end{equation}
where $\mathrm{Drop}$ and $\mathrm{Lin}$ are the dropout and linear functions, respectively; $\mathrm{\sigma}$ is the non-linear activation function (e.g., Relu); $\Theta_0$ and $\Theta_1$ are the trainable parameters. The $\mathrm{MultiGAT}$ is designed as follows:
\begin{equation}
\label{multigat}
\begin{split}
&\mathrm{MultiGAT}(\mathcal{E}, \mathrm{X}_{in}; \mathrm{\Theta}_h) = \mathrm{\Theta}_{h}[\mathrm{head}_1 \, \Vert \, ... \, \Vert \, \mathrm{head}_h],\\
&\mathit{where} \ \mathrm{head}_i = \mathrm{SingleGAT}(\mathcal{E}, \mathrm{X}_{in}; \mathrm{\Theta}_i),
\end{split}
\end{equation}
where $\mathrm{SingleGAT}$ in this paper will be described in detail in Section \ref{subsubsec:singlegat}.

If the $\mathrm{Norm}$ operation is placed before $\mathrm{MultiGAT}$ and $\mathrm{FeedForward}$, then it can be modified as follows:
\begin{equation}
\label{gat-mlp2}
\begin{split}
&\mathrm{X}_\mathrm{gat} = \mathrm{MultiGAT}(\mathcal{E}, \mathrm{Norm}(\mathrm{X}_{in}); \mathrm{\Theta}_\mathrm{gat}) + \mathrm{X}_{in},\\
&\mathrm{X}_{out} = \mathrm{FeedForward}(\mathrm{Norm}(\mathrm{X}_\mathrm{gat}); \mathrm{\Theta}_\mathrm{fed}) + \mathrm{X}_\mathrm{gat}.
\end{split}
\end{equation}

We argue intuitively that unlike the textual and acoustic tasks which rely on sequence-level context modeling, the visual task relies more on the features directly expressed in the current image. A model that is capable of both sequence-level context modeling and feature-level modeling is desired for the multimodal feature fusion. The $\mathrm{MultiGAT}$ sublayer in GAT-MLP can capture sequence-level contextual information, while the $\mathrm{FeedForward}$ sublayer compensates for the failure to efficiently capture feature-level information. Therefore, the combination of the $\mathrm{MultiGAT}$ and $\mathrm{FeedForward}$ in the GAT-MLP layer can mutually compensate for encoding disparities of various modalities in the multimodal task.

\subsubsection{GAT-MLP Based PairCC}\label{subsubsec:paircc}
If the features of multiple modalities are concatenated together directly, then it will not only be challenging to fuse due to heterogeneity gap, but it will also neglect cross-modal interactive information. What's worse, the larger the number of modalities is, the more serious the problem of heterogeneity gap between modalities is. Therefore, we propose the strategy of GAT-MLP based Pair-wise Cross-modal Complementation (PairCC) for cross-modal feature interaction and minimizing the heterogeneity gap. The process of GAT-MLP based PairCC is shown in Fig.~\ref{fig:overall}, which mainly consists of GAT-MLP and concatenation layer. Specifically, we first feed the feature matrices of visual and acoustic modality into GAT-MLP layer for intra-modal context and inter-modal interaction encoding, and concatenate the visual and acoustic encoding results to obtain the $v$-$a$ (visual-acoustic) feature matrix $\mathrm{H}^{va}$; then we treat $\mathrm{H}^{va}$ as the feature matrix of a new modality, and perform the same encoding operation between $\mathrm{H}^{va}$ and the textual feature matrix to obtain the $v$-$a$-$t$ (visual-acoustic-textual) feature matrix $\mathrm{H}^{vat}$; finally, $\mathrm{H}^{vat}$ and the feature matrix of shared subspace are encoded similarly to obtain the final feature matrix. The above steps can be formulated simply as follows:
\begin{equation}
\label{eq:multi_paircc}
\begin{split}
&\mathrm{H}^{va} = \mathrm{PairCC}(\mathcal{E}^{va}, \mathrm{X}_{sep}^{v}, \mathrm{X}_{sep}^{a}; \Theta_{sep}^{va}),\\
&\mathrm{H}^{vat} = \mathrm{PairCC}(\mathcal{E}^{vat}, \mathrm{H}^{va}, \mathrm{X}_{sep}^{t}; \Theta_{sep}^{vat}),\\
&\mathrm{H'} = \mathrm{PairCC}(\mathcal{E}', \mathrm{H}^{vat}, \mathrm{X}^{vat}_{shr}; \Theta'),\\
\end{split}
\end{equation}
where $\mathrm{H'}$ is the final output of feature matrix; $\mathcal{E}^{va}$ denotes the edge set consisting of $\mathcal{E}_{intra}^v$, $\mathcal{E}_{intra}^a$, and $\mathcal{E}_{inter}^{va}$; $\mathcal{E}^{vat}$ and $\mathcal{E}'$ are also the edge sets that are created similarly to $\mathcal{E}^{va}$; $\Theta_{sep}^{va}$, $\Theta_{sep}^{vat}$ and $\Theta'$ are the trainable parameters, respectively; $\mathrm{PairCC}$ indicates GAT-MLP based PairCC function.

\subsubsection{$Single$GAT}\label{subsubsec:singlegat}
We will present the graph attention network of this paper in this part. Graph Neural Networks (GNNs) typically involve two processes: aggregating information with the aggregation function and updating state with the combination function. Following that, we'll describe our $\mathrm{SingleGAT}$ (single-head graph attention) sublayer in terms of the aggregation function $\mathrm{AGG}$ and combination function $\mathrm{COM}$. The two processes can be formalized as follows:
\begin{equation}
\label{eq:agg-com}
\begin{split}
&\mathrm{x}_{agg,i} = \mathrm{AGG}(\{\mathrm{x}_j|{w_j} \in \mathcal{N}(w_i)\};\Theta_{agg}),\\ %,\mathrm{x}_i
&\mathrm{x}_{com,i} = \mathrm{COM}(\mathrm{x}_i, \mathrm{x}_{agg,i}; {\Theta}_{com}), 
\end{split}
\end{equation}
where $\mathrm{AGG}$ and $\mathrm{COM}$ are the aggregation and combination functions, respectively; $\mathrm{x}_i \in \mathrm{X}$ denotes the feature vector of node $w_i$, and $w_i \in \mathcal{V}$; $\mathrm{x}_j$ is the feature vector of $w_i$'s neighbor $w_j$; $\Theta_{agg}$ and ${\Theta}_{com}$ denote the learnable parameters.

\textit{Aggregation:} It is well known that computing the importance of neighbor information is crucial when GNNs aggregate information. So we utilize the attention mechanism to implement the aggregation function $\mathrm{AGG}$. The output of aggregation is expressed as follows:
\begin{equation}
\label{eq:aggregation}
\mathrm{x}_{agg,i} = \sum_{{w_j} \in \mathcal{N}(w_i)} \alpha_{ij} \mathrm{W}_{agg}\mathrm{x}_j,
\end{equation}
where $\alpha_{ij}$ is attention coefficient, as well as the edge weight between node $w_i$ and $w_j$; $w_j$ is the neighboring node of $w_i$; $\mathrm{x}_j$ denotes the feature vector of $w_j$, and $\mathrm{x}_j \in \mathrm{X}$; $\mathrm{W}_{agg}$ denotes the learnable parameter.

\textit{GATv2 based edge weights:} We use the attention module of GATv2~\cite{brody2022how} to learn edge weights for characterizing the relevance of diverse neighbor information. We define the attention coefficient $\alpha_{ij}$ as follows:
\begin{equation}
\label{eq:edge-weights}
\alpha_{ij} =
\frac{
\exp\left(\mathrm{a}^{\top}\sigma\left(\mathrm{\Theta}_\mathrm{att}
[\mathrm{x}_i \, \Vert \, \mathrm{x}_j]
\right)\right)}
{\sum_{w_k \in \mathcal{N}(w_i)}
\exp\left(\mathrm{a}^{\top}\sigma\left(\mathrm{\Theta}_\mathrm{att}
[\mathrm{x}_i \, \Vert \, \mathrm{x}_k]
\right)\right)},
\end{equation}
where $\mathrm{\sigma}$ denotes the non-linear activation function, such as $\mathrm{LeakyReLU}$; $\Vert$ denotes the concatenation operation; $\mathrm{\Theta}_\mathrm{att}$ is the learnable parameter; $\mathrm{x}_i$ is the feature representation of the current node $w_i$; both $\mathrm{x}_j$ and $\mathrm{x}_k$ are the representations of neighboring node of $w_i$. In this work, the neighboring node is either an intra-modal contextual node or an inter-modal interactive node of $w_i$.

\textit{Embedding of edge types:} We assume that different types of edge/relation involve different implicit dependency information in the dialogue. Here are two conjectures:
\begin{itemize}
\item[\ding{172}] Suppose $w_j$ is an intra-modal contextual neighbor node of $w_i$ ($w_j$ is a long-distance contextual node). $w_j$ and $w_i$ may have similar semantics when they are uttered by the same speaker. At this moment, $w_j$ is more critical relative to others that have different speakers from $w_i$. 
\item[\ding{173}] Suppose $w_j$ is the inter-modal interactive neighbor node of $w_i$. When the semantics of $w_i$ does not match the ground-truth emotion label, $w_j$ can be semantically complementary to $w_i$. $w_j$ is more important at this time relative to other neighbor nodes.
\end{itemize}
 
Therefore, we encode the edge types as vector representations, and put them into the attention module to aid in the computation of the attention coefficient. We consider that the edge weight is affected not only by the nodes, but also by the edge types. The embedding of edge types, i.e., feature of edge type, can be formalized as follows:
\begin{equation}
\label{eq:embed-edge-types}
\begin{split}
&ET_{emb} = \mathrm{Embedding}(\mathrm{ET}, DM),\\
&\mathit{where} \ DM =  M \times (D^2  + D + M-1)/2,
\end{split} 
\end{equation}
where $\mathrm{ET} = \mathrm{ET}_{intra} \cup \mathrm{ET}_{inter}$ denotes the set of edge types, and $DM$ is the number of edge types in a dialogue with $D$ speakers and $M$ modalities. The attention coefficient with the addition of edge feature is computed as follows:
\begin{equation}
\label{eq:edge-weights2}
\alpha'_{ij} =
\frac{
\exp\left(\mathrm{a}^{\top}\sigma\left(\mathrm{\Theta}_\mathrm{att}
[\mathrm{x}_i \, \Vert \, \mathrm{x}_j \, \Vert \, \mathrm{et}_{ij}]
\right)\right)}
{\sum_{w_k \in \mathcal{N}(w_i)}
\exp\left(\mathrm{a}^{\top}\mathrm{\sigma}\left(\mathrm{\Theta}_\mathrm{att}
[\mathrm{x}_i \, \Vert \, \mathrm{x}_k \, \Vert \, \mathrm{et}_{ik}]
\right)\right)},
\end{equation}
where $\mathrm{et}_{ij} \in ET_{emb}$ denotes the edge feature of between utterance node $w_i$ and $w_j$.

\textit{Combination:} The combination function $\mathrm{COM}$ combines $\mathrm{x}_{agg,i}$ with $\mathrm{x}_i$. We employ GRU as the combination function, which is inspired by GraphSage~\cite{hamilton2017inductive} but different from it. The output of the graph attention is expressed as follows:
\begin{equation}
\label{eq:combination}
\mathrm{x}_{com,i}^{fwd} = \mathrm{GRU}(\mathrm{x}_i, \mathrm{x}_{agg,i}; {\Theta}_{com}^{fwd}), 
\end{equation}
where $\mathrm{x}_{com,i}^{fwd}$, $\mathrm{x}_i$ and $\mathrm{x}_{agg,i}$ are the output, input and hidden state of GRU, respectively; ${\Theta}_{com}^{fwd}$ is the trainable parameter. The neighbor information $\mathrm{x}_{agg,i}$ (including intra-modal contextual information and inter-modal interactive information) is employed as the hidden state of GRU, and it may not be completely exploited. Therefore, we reverse the order of $\mathrm{x}_i$ and $\mathrm{x}_{agg,i}$, i.e., $\mathrm{x}_{agg,i}$ and $\mathrm{x}_i$ are respectively utilized as the input and hidden state of GRU:
\begin{equation}
\label{eq:combination2}
\mathrm{{x}}_{com,i}^{rev} = \mathrm{GRU}(\mathrm{x}_{agg,i}, \mathrm{x}_i; {\Theta}_{com}^{rev}),
\end{equation}
where $\mathrm{{x}}_{com,i}^{rev}$ is the output, input and hidden state of GRU, and ${\Theta}_{com}^{rev}$ is the trainable parameter. The final output of the single-head graph attention $\mathrm{SingleGAT}$ as follows:
\begin{equation}
\label{eq:singlegat}
\mathrm{x}_{com,i} = \mathrm{{x}}_{com,i}^{fwd} + \mathrm{x}_{com,i}^{rev}.
\end{equation}
By calculating the average of multiple single-head graph attentions, we can obtain the following result:
\begin{equation}
\label{eq:gat}
\mathrm{x}_{\mathrm{gat},i} = \frac{1}{K}  \sum_{k = 1}^{K} \mathrm{x}_{com,i}^k,
\end{equation}
where $\mathrm{x}_{\mathrm{gat},i} \in \mathrm{X}_\mathrm{gat} $ is the output of the multi-head graph attention network, and $K$ denotes the number of heads.

\subsection{Multimodal Emotion Classifier}
After encoding with the GAT-MLP based PairCC, the feature vector $h_i \in H'$ of utterance $u_i$ can be obtained. It is then fed to the fully connected layer to predict the emotion label $\hat{y}_i$ for the utterance $u_i$:
\begin{equation}
\label{eq:predict}
\begin{split}
v_i &= \mathrm{ReLU}(\mathrm{W}_0h_i+\mathrm{b}_0),\\
p_i &= \mathrm{Softmax}(\mathrm{W}_1v_i+\mathrm{b}_1),\\
\hat{y}_i &= \mathop{\mathrm{argmax}}_\mathrm{k}(p_i[\mathrm{k}]),
\end{split}
\end{equation}

We employ cross-entropy loss along with L2-regularization as classification loss function to train the model:
\begin{equation}
\label{eq:class-loss}
\mathcal{L}_{cls} = - \frac {1}{\sum_{k=0}^{N-1} n(k)} \sum_{i=0}^{N-1}\sum_{j=0}^{n(i)-1} y_{ij} \log p_{ij} + \lambda \lvert \mathrm{\Theta}_{re} \rvert,
\end{equation}
where $N$ is the number of dialogues, $n(i)$ is the number of utterances in dialogue $i$; $y_{ij}$ denotes the ground truth label of the $j$-th utterance in the $i$-th dialogue, $p_{ij}$ denotes the probability distribution of predicted emotion label of the $j$-th utterance in the $i$-th dialogue; $\lambda$ is the L2-regularizer weight, and $\mathrm{\Theta}_{re}$ is the trainable parameter. 

Finally, combining the shared subspace loss $\mathcal{L}_{shr}$, separate subspace losses $\mathcal{L}_{sep}^{\zeta}$ (${\zeta} \in \{a,v,t\}$) and classification loss $\mathcal{L}_{cls}$ together, the final objective function is computed as:
\begin{equation}
\label{eq:loss}
\mathcal{L} = \mathcal{L}_{cls} + \beta\mathcal{L}_{shr} + \gamma^a\mathcal{L}_{sep}^a + \gamma^v\mathcal{L}_{sep}^v + \gamma^t\mathcal{L}_{sep}^t,
\end{equation}
where $\beta$, $\gamma^a$, $\gamma^v$, $\gamma^t$ are the trade-off parameters.

\section{Experiment}\label{sec:experiment}
\subsection{Datasets and Evaluation Metrics}
\subsubsection{Datasets}
We evaluate our GraphCFC model on two multimodal benchmark datasets: IEMOCAP~\cite{busso2008iemocap} and MELD~\cite{poria2018meld}, which are subjected to raw utterance-level feature extraction according to MMGCN~\cite{hu2021mmgcn}. The statistics of them are shown in TABLE~\ref{tab:statistic}. 
\begin{table}[htbp]
\centering
\renewcommand{\arraystretch}{1.0}
\setlength{\tabcolsep}{3pt}
\caption{The statistics of IEMOCAP and MELD}
\begin{tabular}{c|ccc|ccc|c|c}
\hline
\multirow{2}[0]{*}{Dataset} & \multicolumn{3}{c|}{Dialogues} & \multicolumn{3}{c|}{Uterances} & \multirow{2}[0]{*}{Classes} & \multirow{2}[0]{*}{\makecell{Speakers in\\a Dialogue}} \\  
        & train & valid & test & train & valid & test &       &     \\
\hline
IEMOCAP & \multicolumn{2}{c}{120} & 31    & \multicolumn{2}{c}{5810} & 1623  & 6     & 2     \\
MELD  & 1039  & 114   & 280   & 9989  & 1109  & 2610  & 7     & 3 or more \\
\hline
\end{tabular}%
\label{tab:statistic}%
\end{table}%
      
\textbf{IEMOCAP} is a multimodal dataset of two-way conversations from ten professional actors. It contains 151 conversations, a total of 7433 dyadic utterances. Emotion labels of IEMOCAP include \textit{Neutral}, \textit{Happy}, \textit{Sad}, \textit{Angry}, \textit{Frustrated} and \textit{Excited}. As in previous works~\cite{ghosal2019dialoguegcn, hu2021mmgcn}, we utilize the first 80\% of the data as the training set and the remaining data as the test set, with the 10\% of the training set used as the validation set. IEMOCAP is one of the most popular datasets in ERC task, with high quality and multimodal information.

\textbf{MELD} is a multimodal dataset, containing videos of multi-party conversations from Friends TV series. It involvs over 1433 conversations, a total of 13708 utterances by 304 speakers. Distinct from IEMOCAP, each conversation in MELD includes three or more speakers. Emotion labels include \textit{Anger}, \textit{Disgust}, \textit{Sadness}, \textit{Joy}, \textit{Neutral}, \textit{Surprise} and \textit{Fear}. The conversations in this dataset involve many backgrounds knowledge, which makes it challenging to recognize the right emotion.

\subsubsection{Metrics}
Following the previous methods~\cite{ghosal2019dialoguegcn, hu2021mmgcn}, we chose weighted-average F1 score as the evaluation metric due to the class imbalanced problem. F1 score is reported for each class to allow for a more comprehensive comparison with the baselines. We also record the average accuracy score in addition to the weighted-average F1 score.

\subsection{Baselines}
To verify the effectiveness of our proposed GraphCFC model, we compare it with several previous baselines. The baselines include bc-LSTM~\cite{poria2017context}, CMN~\cite{hazarika2018conversational}, ICON~\cite{hazarika2018icon}, DialogueRNN~\cite{majumder2019dialoguernn}, DialogueGCN~\cite{ghosal2019dialoguegcn}, DialogueCRN~\cite{hu2021dialoguecrn} and MMGCN~\cite{hu2021mmgcn}. The details of these models are listed as follows.

\textbf{bc-LSTM} encodes context-aware information through a bidirectional LSTM network, but without taking speaker-related information into account. 
\textbf{CMN} models utterance context through speaker-dependency GRUs, but it can only work when the conversation includes two speakers. \textbf{ICON} has improved CMN by modeling distinct speakers. A global GRU is utilized to model the variance of emotion status in a conversation. Nevertheless, ICON still can't be applied in scenario with more than two speakers. \textbf{DialogueRNN} leverages three GRUs to model information of speakers and sequence in conversations, which contain Global GRU, Speaker GRU and Emotion GRU. The goals of three GRUs are to extract context information, model identity information of the speakers and detect emotion of utterances, respectively. \textbf{DialogueGCN} focuses on the function of GCN, i.e., aggregating neighbor information, to improve the performance of ERC tasks. We extend DialogueGCN by directly concatenating features of each modality to implement multimodal setting. \textbf{DialogueCRN} extracts and integrates emotional clues by devising multi-turn reasoning modules to sufficiently model the situation-level and speaker-level context in a conversation. In order to achieve multimodal setting, we concatenate features of three modalities simply. \textbf{MMGCN} adopts a graph-based approach for multimodal feature fusion. MMGCN is currently significantly superior to most baselines for multimodal ERC, which provides a new idea for multimodal fusion.

\subsection{Implementation Details}
We implement the GraphCFC model through the PyTorch framework, and all experiments are executed on NVIDIA Tesla A100. The optimizer is AdamW, the L2 regularization parameter is 0.00001, and the Dropout rate is 0.1. For IEMOCAP dataset, the number of GAT-MLP layers is 5, the learning rate is 0.00001, the ratio of the speaker embedding $\mu$ is 1.0, and the batch size is 8. For MELD dataset, the number of GAT-MLP layers is 3, the learning rate is 0.00001, the ratio of the speaker embedding $\mu$ is 0.7, and the batch size is 32. We utilize the method proposed by Kendall et al.~\cite{kendall2018multi} to set the trade-off parameters ($\beta$, $\gamma^a$, $\gamma^v$ and $\gamma^t$) of multiple loss functions as learnable parameters instead of setting them manually.

\section{Results and Analysis}\label{sec:results_analysis}
In this section, we report and discuss the results of all comparative experiments and ablation studies. In addition, we provide three case studies on the IEMOCAP dataset at the end of this section.

\subsection{Overall Performance}
\begin{table*}[htbp]
\centering
\renewcommand{\arraystretch}{1.0}
\setlength{\tabcolsep}{3pt}
\caption{The Overall Performance of All Models on Both IEMOCAP and MELD Datasets Under the Multimodal Setting}
\begin{threeparttable}
\begin{tabular}{c|cccccc|cc||ccccc|cc}
\hline
\multicolumn{1}{c|}{\multirow{2}{*}{Model}} & \multicolumn{8}{c||}{IEMOCAP} & \multicolumn{7}{c}{MELD} \\\cline{2-16}          
& \textit{Happy} & \textit{Sad}   & \textit{Neutral} & \textit{Angry} & \textit{Excited} & \textit{Frustrated} & Accuracy & wa-F1 & \textit{Neutral} & \textit{Surprise} & \textit{Sadness} & \textit{Joy} & \textit{Anger} & Accuracy & wa-F1 \\ \hline
bc-LSTM & 32.63 & 70.34 & 51.14 & 63.44 & 67.91 & 61.06 & 59.58 & 59.10  & 75.66 & 48.47 & 22.06 & 52.10  & 44.39 & 59.62 & 56.80 \\
CMN   & 30.38 & 62.41 & 52.39 & 59.83 & 60.25 & 60.69 & 56.56 & 56.13 &  - & - & - & - & - & - & - \\
ICON  & 29.91 & 64.57 & 57.38 & 63.04 & 63.42 & 60.81 & 59.09 & 58.54 &  - & - & - & - & - & - & - \\
%   MulT  &  &  &  &  &  &  &  &  &  - & - & - & - & - & - & - \\
DialogueRNN & 33.18 & 78.80  & 59.21 & 65.28 & 71.86 & 58.91 & 63.40  & 62.75 & 76.79 & 47.69 & 20.41 & 50.92 & 45.52 & 60.31 & 57.66 \\
DialogueCRN & \textbf{51.59} & 74.54 & 62.38 & 67.25 & 73.96 & 59.97 & 65.31 & 65.34 & 76.13 & 46.55 & 11.43 & 49.47 & 44.92 & 59.66 & 56.76 \\
DialogueGCN & 47.10  & 80.88 & 58.71 & 66.08 & 70.97 & 61.21 & 65.54 & 65.04 & 75.97 & 46.05 & 19.60  & 51.20  & 40.83 & 58.62 & 56.36 \\
MMGCN & 45.45 & 77.53 & 61.99 & 66.67 & 72.04 & \textbf{64.12} & 65.56 & 65.71 & 75.16 & 48.45 & 25.71 & \textbf{54.41} & 45.45 & 59.31 & 57.82 \\
\hline
GraphCFC  & 43.08 & \textbf{84.99} & \textbf{64.70} & \textbf{71.35}  & \textbf{78.86} & 63.70 & \textbf{69.13} & \textbf{68.91} & \textbf{76.98} & \textbf{49.36} & \textbf{26.89} & 51.88 & \textbf{47.59} & \textbf{61.42} & \textbf{58.86} \\
\hline
\end{tabular}%
\label{tab:overall7}%
% \begin{tablenotes}
IEMOCAP includes 6 labels, and MELD includes 7 labels (2 of them, i.e., \textit{Fear} and \textit{Disgust}, are not reported due to their statistically insignificant results). Evaluation metrics contain accuracy score ($\mathrm{Accuracy}$ (\%)) and weighted-average F1 score ($\mathrm{wa}$-$\mathrm{F1} $ (\%)). Best performances are highlighted in bold.
% \end{tablenotes}
\end{threeparttable}
\end{table*}%
We compare our proposed GraphCFC with the baseline models on the IEMOCAP and MELD datasets. The overall performance of all models is shown in TABLE~\ref{tab:overall7}. Based on the experimental findings, we can see that the accuracy and weighted-average F1 score of our proposed model is 3.57\% and 3.20\% higher than those of the best baseline model (i.e., MMGCN) on the IEMOCAP dataset. The GraphCFC model achieves higher F1 scores than MMGCN in the most emotions when each emotion is observed separately. The F1 scores of \textit{Sad} and \textit{Excited} are respectively 84.99\% and 78.86\% on the IEMOCAP dataset, which are higher than those of other emotions. For \textit{Sad} and \textit{Excited} emotions, the F1 scores of GraphCFC are far higher than those of MMGCN, which are 7.46\% and 6.82\% higher than those of MMGCN, respectively. Overall, the GraphCFC model outperforms the others in terms of accuracy and weighted-average F1 scores. Therefore, we can conclude that our method can more adequately extract long-distance intra-modal contextual information and inter-modal interactive information in comparison to the baselines such as MMGCN.

We note that while DialogueCRN can achieve excellent performance in uni-modal setting~\cite{hu2021dialoguecrn}, direct concatenation of the results from multiple modalities is not as effective. One probable reason is that direct concatenation generates redundant information and fails to capture the inter-modal interactive information. GraphCFC, in contrast, extracts the interactive information through a GNN-based approach while also reducing redundant information, resulting in superior performances.

Although GraphCFC model outperforms other approaches on the MELD dataset, its improvement was not very significant. It can be observed that the improvements in accuracy and weighted-average F1 scores of GraphCFC are 2.11\% and 1.04\%, respectively, relative to those of MMGCN. The reason for these results may be that the utterance sequences of a dialogue on the MELD dataset aren't from a continuous conversation in real scenes. Therefore, the graph-based models do not take advantage of their ability to capture contextual information. Another potential reason is that the MELD dataset contains a lot of background noise that is unrelated to emotion due to the camera setup and recording conditions. In addition, we observe that the F1 score of the \textit{Sadness} class is low in the results of the MELD dataset. By looking at the distribution of classes in the MELD dataset, we find that the dataset suffers from the class-imbalanced problem. And the \textit{Sadness} class belongs to the minority class, thus resulting in its low F1 score.

\subsection{Comparison Under Various Modality Settings}
\begin{table}[htbp]
\centering
\renewcommand{\arraystretch}{1.0}
\setlength{\tabcolsep}{8pt}
\caption{The Performance of GraphCFC Under Various Modality Setting}
\begin{threeparttable}
\begin{tabular}{c|cc||cc}
\hline
\multicolumn{1}{c|}{\multirow{2}[1]{*}{Modality Setting}} & \multicolumn{2}{c||}{IEMOCAP}  & \multicolumn{2}{c}{MELD} \\
\cline{2-5}
& Accuracy & wa-F1 & Accuracy & wa-F1 \\
\hline
A     & 54.16  & 53.85  & 47.55  & 41.62  \\
V     & 31.61  & 27.67  & 47.59  & 33.26  \\
T     & 59.95  & 60.09  & 60.77  & 56.81  \\
\hline
A + V   & 54.17  & 53.89  & 47.61  & 41.67  \\
A + T   & 64.20  & 64.74  & 59.96  & 57.46  \\
V + T   & 63.15  & 62.96  & 59.46  & 57.29  \\
\hline
A + V + T & \textbf{69.13}  & \textbf{68.91}  & \textbf{61.42} & \textbf{58.86} \\
\hline
\end{tabular}%
\label{tab:modality-setting}%
A, V, and T indicate acoustic, visual, and textual modalities, respectively. Note that shared space mapping and $\mathcal{E}_{inter}$ is not performed in uni-modal settings.
\end{threeparttable}
\end{table}%
TABLE~\ref{tab:modality-setting} shows the performance of GraphCFC model in different modality settings. Overall, the performance of multimodal settings outperforms that of uni-modal settings. The textual modality has the best performance in the uni-modal settings, whereas the visual modality has the lowest result. One probable explanation for the poor result is that the visual modality contains a lot of noise due to the effects of camera position, background, lighting, etc. The combination of textual and acoustic modalities produces the best performance in the two-modal settings, whereas the combination of visual and acoustic modalities produces the worst result. As expected, the fusion of acoustic, visual and textual modalities can improve the performance of GraphCFC.
      
\subsection{Effect of Various Components in GAT-MLP Layer}
\begin{table}[htbp]
\centering
\renewcommand{\arraystretch}{1.0}
\setlength{\tabcolsep}{5pt}
\caption{The Results of GraphCFC When $\mathrm{MultiGAT}$ or $\mathrm{FeedForward}$ is Not Used in the GAT-MLP Layer}
\begin{threeparttable}
\begin{tabular}{c|c|cc||cc}
\hline
\multicolumn{1}{c|}{\multirow{2}[0]{*}{MultiGAT}} &\multicolumn{1}{c|}{\multirow{2}[0]{*}{FeedForward}} & \multicolumn{2}{c||}{IEMOCAP} & \multicolumn{2}{c}{MELD} \\
\cline{3-6}
& & Accuracy & wa-F1 & Accuracy & wa-F1 \\
\hline
- w/o & - w & 64.20  & 64.55  & 60.50  & 58.03  \\
- w & - w/o &67.04  & 67.24  & 60.84  & 58.43  \\
%   w/o GAT w/o FeedForward & 64.02  & 64.27  & 60.61  & 57.95  \\
\hline
- w & - w &\textbf{69.13}  & \textbf{68.91}  & \textbf{61.42}  & \textbf{58.86}  \\
\hline
\end{tabular}%
\label{tab:components}%
$\mathrm{w/o}$ and $\mathrm{w}$ are respectively $\mathrm{without}$ and $\mathrm{with}$, where $\mathrm{w/o}$ ($\mathrm{w}$) indicates the non-use (use) of $\mathrm{MultiGAT}$ or $\mathrm{FeedForward}$ sublayer.
\end{threeparttable}
\end{table}%
\begin{figure}[htbp]
\centering
\includegraphics[width=3.4in]{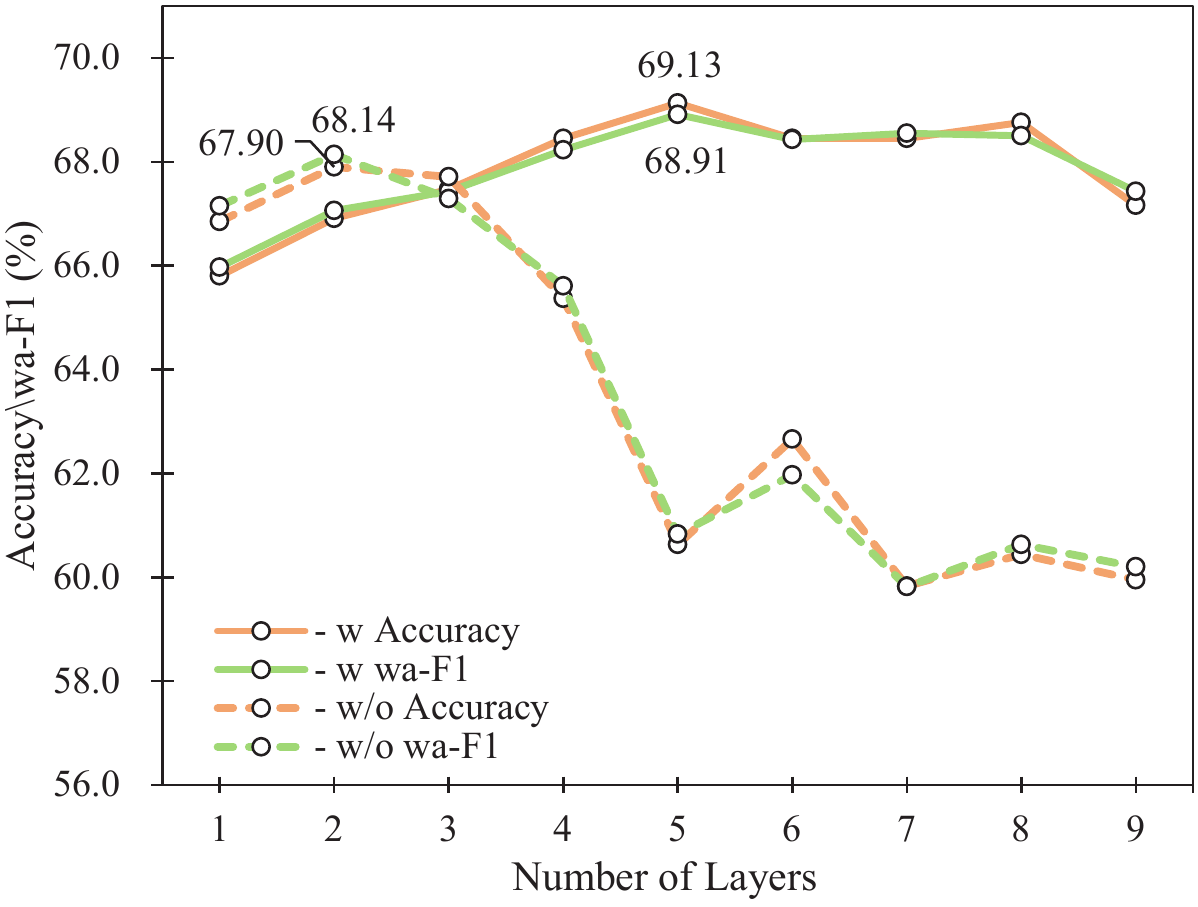}
\caption{The effects of the number of GAT-MLP layer and skip connection on our model. The figure shows the results on the IEMOCAP dataset. $\mathrm{w}$ ($\mathrm{w/o}$) indicates the use (non-use) of skip connection.}
\label{fig:skip}
\end{figure}
We report the effects of $\mathrm{MultiGAT}$ and $\mathrm{FeedForward}$ in the GAT-MLP layer in TABLE~\ref{tab:components}. The performance of our model is noticeably degraded when the $\mathrm{MultiGAT}$ or $\mathrm{FeedForward}$ sublayer is not adopted. When the $\mathrm{MultiGAT}$ sublayer is not utilized on the IEMOCAP dataset, the accuracy and weighted-average F1 scores of GraphCFC decrease by 4.93\% and 4.36\%, respectively. The accuracy and F1 scores respectively decrease by 2.09\% and 1.67\% when the $\mathrm{FeedForward}$ sublayer is not applied. As a result, we can deduce that the effect of $\mathrm{MultiGAT}$ in the GAT-MLP layer is more significant than that of $\mathrm{FeedForward}$.

The effects of different numbers of GAT-MLP layer and skip connection on the GraphCFC model are shown in Fig.~\ref{fig:skip}. We can see that if we remove the skip connection, the performance of the model will drop sharply on the IEMOCAP dataset as the number of GAT-MLP layer increases when a certain threshold is exceeded. On the contrary, if we keep the skip connection, the performance of the proposed GraphCFC decreases slowly. Therefore, skip connection can help to mitigate the problem of over-smoothing to a certain extent.

\subsection{Effect of Multi-Subspace Loss}
The impacts of the multi-subspace loss functions are seen in TABLE~\ref{tab:muti-subspace}. After eliminating the shared subspace loss or separate subspace loss, both accuracy and weighted-average F1 scores decline, as seen in the TABLE~\ref{tab:muti-subspace}. The experimental results suggest that setting the loss function in the multi-subspace extractor can effectively improve the performance of our GraphCFC.
\begin{table}[htbp]
\centering
\renewcommand{\arraystretch}{1.0}
\setlength{\tabcolsep}{7pt}
\caption{The Effect of the Multi-Subspace Loss Functions}
\begin{threeparttable}
\begin{tabular}{c|c|cc||cc}
\hline
\multicolumn{1}{c|}{\multirow{2}[0]{*}{$\mathcal{L}_{shr}$}} & \multicolumn{1}{c|}{\multirow{2}[0]{*}{$\mathcal{L}_{sep}^{\zeta}$}}& \multicolumn{2}{c||}{IEMOCAP} & \multicolumn{2}{c}{MELD} \\
\cline{3-6}
& & Accuracy & wa-F1 & Accuracy & wa-F1 \\
\hline
- w/o & - w & 68.70  & 68.35  & 61.00  & 58.39  \\
- w & - w/o & 67.53  & 67.56  & 60.27  & 57.99  \\
- w/o & - w/o & 68.70  & 68.36  & 60.38  & 58.09  \\
\hline
- w & - w & \textbf{69.13}  & \textbf{68.91}  & \textbf{61.42}  & \textbf{58.86}  \\
\hline
\end{tabular}%
\label{tab:muti-subspace}%
$\mathcal{L}_{shr}$ and $\mathcal{L}_{sep}^{\zeta}$ (${\zeta} \in \{a,v,t\}$) denote shared and separate subspace losses, respectively.
\end{threeparttable}
\end{table}%
      
\subsection{Effect of Multi-speaker and Edge Types}
\begin{table}[htbp]
\centering
\renewcommand{\arraystretch}{1.0}
\setlength{\tabcolsep}{7pt}
\caption{The Influence of Speakers and Edge Types on Our GraphCFC Model}
\begin{threeparttable}
\begin{tabular}{c|c|cc||cc}
\hline
\multicolumn{1}{c|}{\multirow{2}[0]{*}{${S_{emb}}$}}& \multicolumn{1}{c|}{\multirow{2}[0]{*}{$E_{emb}$}} & \multicolumn{2}{c||}{IEMOCAP} & \multicolumn{2}{c}{MELD} \\
\cline{3-6}
& & Accuracy & wa-F1 & Accuracy & wa-F1 \\
\hline
- w/o   & - w & 68.02  & 68.04  & 60.69  & 58.35  \\
- w  & - w/o & 65.26  & 65.91  & 60.46  & 57.91  \\
\hline
- w   & - w &\textbf{69.13}  & \textbf{68.91}  & \textbf{61.42}  & \textbf{58.86}  \\
\hline
\end{tabular}%
\label{tab:speaker-edgetype}%
${S_{emb}}$ and $E_{emb}$ indicate the embeddings of multi-speaker and edge types, respectively.
\end{threeparttable}
\end{table}%
The influence of speakers and edge types on our GraphCFC model is seen in TABLE~\ref{tab:speaker-edgetype}. The performance of GraphCFC will be compromised if the embedding of multi-speaker or edge types is not employed. The weighted-average F1 score drops to 65.91\% on the IEMOCAP dataset when the embedding of edge types is not utilized, which amply proves our hypothesis that edge types affects the relevance of neighbor information. We note that without adding speaker information, the results of GraphCFC show only a slight degradation, which is still higher than the results of baseline models. The phenomenon demonstrates that GrpahCFC is not heavily dependent on speaker and has a certain degree of generalization capability. Generally speaking, the performance of our proposed method can be improved by adding the embeddings of multi-speaker and edge types.

\subsection{Effect of the Past $j$ and Future $k$ Utterance Nodes}
\begin{figure}[htbp]
\centering
\includegraphics[width=3.4in]{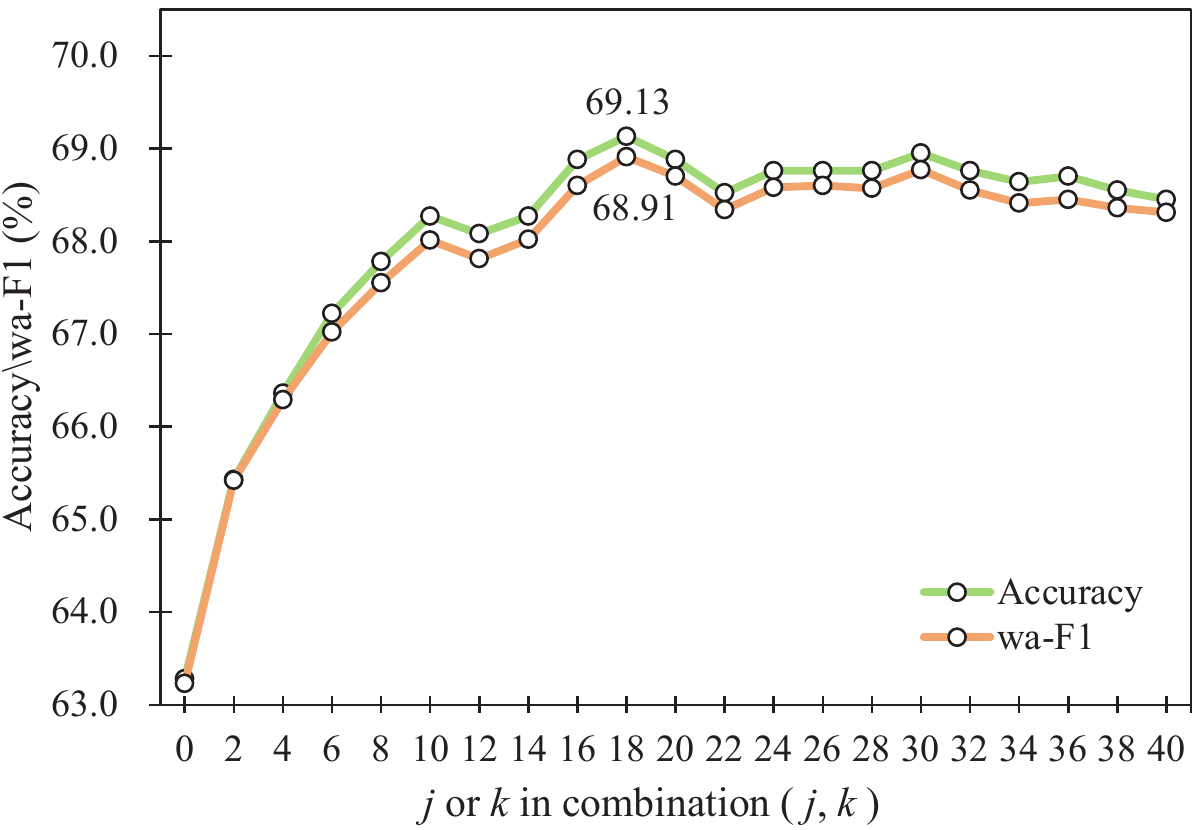}
\caption{The effects of $j$ nodes in the past and $k$ nodes in the future on the proposed GraphCFC model. The figure shows the results on the IEMOCAP dataset.}
\label{fig:past_future}
\end{figure}

As shown in Fig.~\ref{fig:past_future}, we discuss the effect of past $j$ nodes and future $k$ nodes on our proposed GraphCFC model. We set $j$ and $k$ to multiple combinations (the combination can be denoted as $(j, k)$), such as $(0,0)$, $(2,2)$, $(4, 4)$, $(6, 6)$, ..., $(40, 40)$. From Fig.~\ref{fig:past_future}, it can be concluded that the accuracy and weighted-average F1 scores increase on the IEMOCAP dataset with increasing values of $j$ and $k$. When a certain threshold combination (i.e., $(j, k) = (18, 18)$) are reached, however, the accuracy and F1 scores gradually decrease. In particular, GraphCFC performs worst when the conversational context is not available (i.e., setting both $j$ and $k$ set to 0). Therefore, we can draw the conclusion that the conversational context is a crucial parameter for the proposed method.

\subsection{Overall Performance of Three-Emotion}
In this part, we conduct comparative experiments of the three-emotion. Prior to model training, we merge the original emotion labels into three categories (i.e., \textit{Positive}, \textit{Neutral}, and \textit{Negative}), while the proposed GraphCFC is transformed into a three-classification model. Specifically, TABLE~\ref{tab:three-emotion} shows the statistics of the merged emotion labels.
\begin{table}[htbp]
\centering
\renewcommand{\arraystretch}{1.0}
% \small
\setlength{\tabcolsep}{6pt}
\caption{The Statistics of the Merged Emotion Labels}
\begin{tabular}{c|c|c}
\hline
New Label & IEMOCAP & MELD \\
\hline
\textit{Positive} & \textit{Happy}, \textit{Excited} & \textit{Joy} \\
\textit{Negative} & \textit{Sad}, \textit{Angry}, \textit{Frustrated} & {\makecell[c]{\textit{Surprise}, \textit{Fear}, \textit{Sadness},\\ \textit{Disgust}, \textit{Anger}}} \\
\textit{Neutral} & \textit{Neutral} & \textit{Neutral} \\
\hline
\end{tabular}
\label{tab:three-emotion}
\end{table}

\begin{table*}[hbtp]
\centering
\renewcommand{\arraystretch}{1.0}
% \small
\setlength{\tabcolsep}{5pt}
\caption{The Overall Performance After Converting the Dataset Into Three-Emotion Labels Under the Multimodal Setting}
\begin{tabular}{c|ccc|cc||ccc|cc}
\hline
\multicolumn{1}{c|}{\multirow{2}{*}{Model}} & \multicolumn{5}{c||}{IEMOCAP} & \multicolumn{5}{c}{MELD} \\
\cline{2-11}
& \textit{Positive} & \textit{Neutral} & \textit{Negative} & Accuracy & wa-F1 & \textit{Positive} & \textit{Neutral} & \textit{Negative} & Accuracy & wa-F1 \\
\hline
bc-LSTM & \textbf{90.58} & 55.63 & 84.04 & 79.54 & 79.10  & 36.97 & 75.12 & 61.46 & 65.13 & 64.26 \\
DialogueRNN & 88.36 & 57.99 & 83.81 & 78.87 & 78.94 & 40.29 & 74.95 & 62.10  & 65.52 & 64.93 \\
DialogueCRN & 79.39 & 61.51 & 83.09 & 75.66 & 76.97 & 40.80  & 74.40  & 62.87 & 65.98 & 65.32 \\
DialogueGCN & 84.22 & 56.88 & 83.66 & 77.57 & 77.48 & 32.92 & \textbf{75.64} & 63.96 & 66.67 & 64.80 \\
MMGCN & 85.20  & \textbf{64.21} & 83.73 & 79.36 & 79.95 & 43.32 & 75.5  & 65.57 & 67.93 & 66.92 \\
\hline
GraphCFC  & 88.48 & 62.03 & \textbf{84.35} & \textbf{79.91} & \textbf{80.20}  & \textbf{50.66} & 75.12 & \textbf{66.26} & \textbf{68.54} & \textbf{68.12} \\
\hline
\end{tabular}
\label{tab:overall3}
\end{table*}
The experimental results of our three-emotion are recorded in TABLE~\ref{tab:overall3}. We can find that the experimental results are similar to those of the previous experiments with six or seven-emotion, with improved accuracy and weighted-average F1 scores for all models. It can be seen that the accuracy and F1 scores of GraphCFC improve 0.61\% and 1.20\% relative to those of MMGCN on the MELD dataset, respectively. Similarly, there are slight improvements of accuracy and F1 scores on the IEMOCAP dataset. It may be attributable to the fact that the emotion labels are coarsened (similar emotions like \textit{Excited} and \textit{Happy} are merged) after converting the dataset into three-emotion labels, so most of the models are capable of performing the task of emotion classification easily.

\subsection{Case Studies}
\begin{figure*}[hbtp]
\centering
\subfloat[]{\includegraphics[width=1.805in]{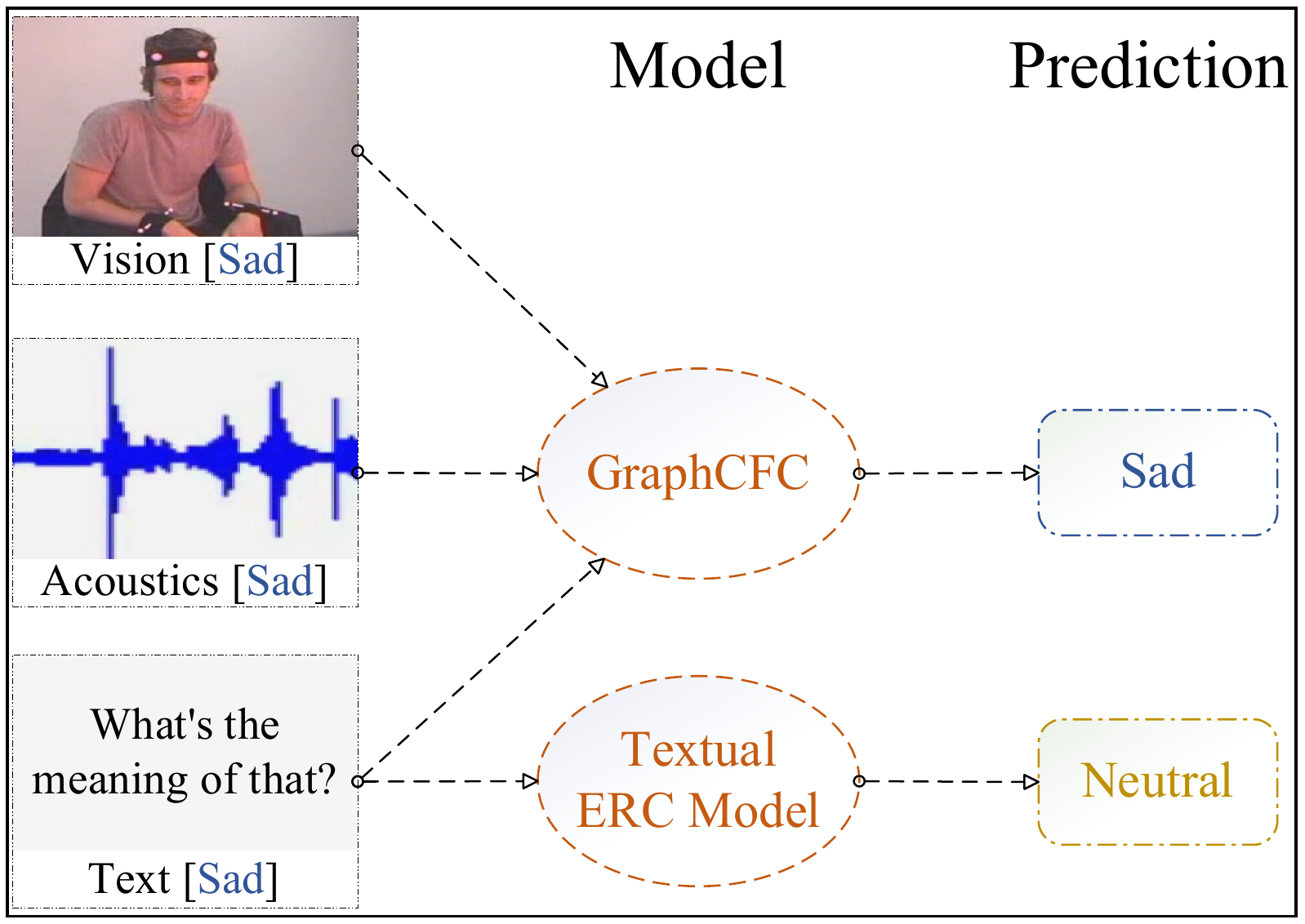}%
\label{fig:case1}}
\hfil
\subfloat[]{\includegraphics[width=2.75in]{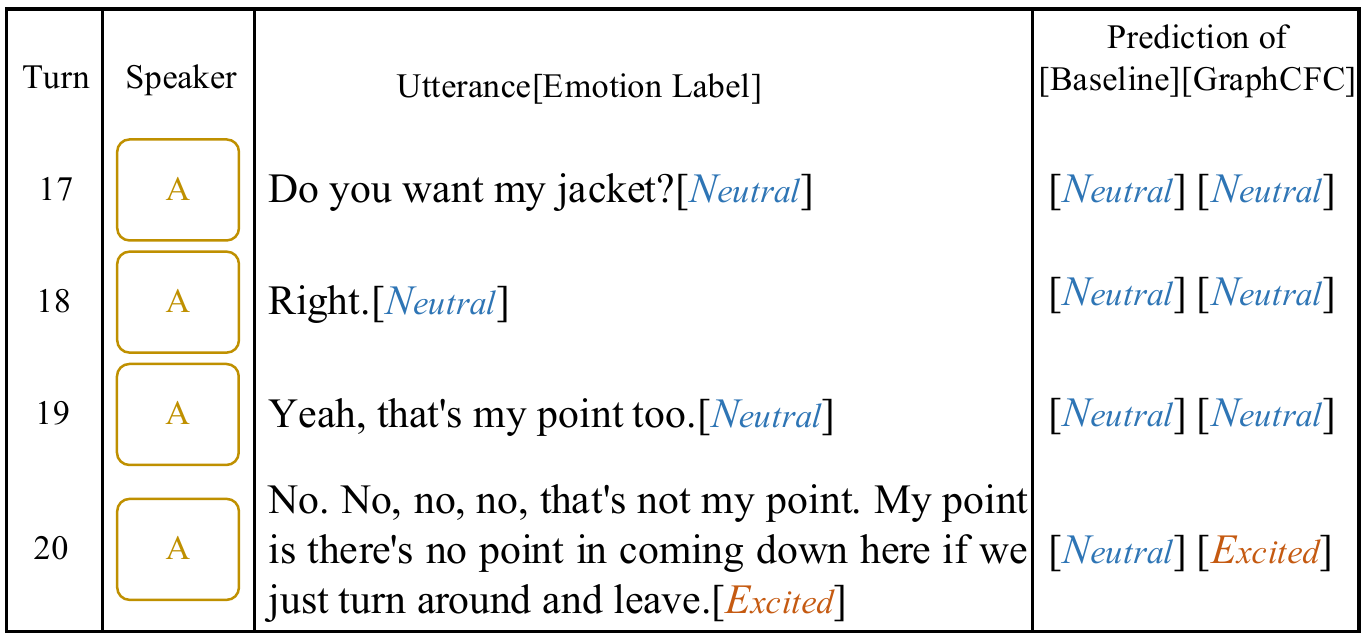}%
\label{fig:case2a}}
\hfil
\subfloat[]{\includegraphics[width=2.3in]{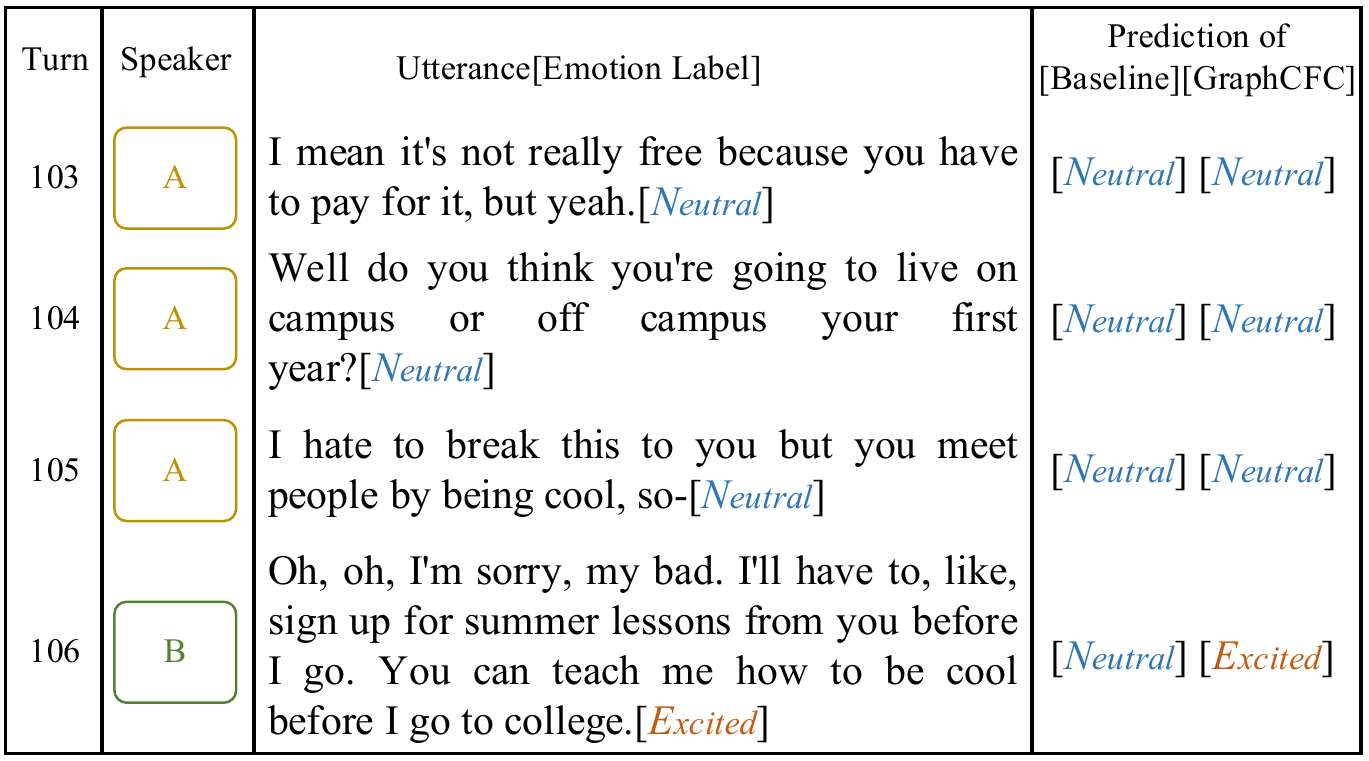}%
\label{fig:case2b}}
\caption{The cases of ERC on the IEMOCAP. (a) An example shows that multi-modality can be used to compensate for the shortcoming of single-textual modality. (b) Emotional-shift in one-speaker scenario. (c) Emotional-shift in two-speaker scenario.}
\label{fig:case}
\end{figure*}

As shown in Fig.~\ref{fig:case}, we conduct several case studies in this part. In text-modal ERC models such as DialogueGCN and DialogueCRN, several utterances with non-\textit{Neutral} emotion labels, such as ``\textbf{okay.}", ``\textbf{yeah.}" or ``\textbf{What's the meaning of that?}", are usually recognized as \textit{Neutral} directly. In contrast, multimodal ERC models such as GraphCFC make integrated judgments based on multiple modalities, which, for example, may eventually be recognized as \textit{Sad}. Therefore, visual and acoustic modalities can compensate for such lapses. Fig.~\ref{fig:case1} depicts the above-mentioned case on the IEMOCAP dataset.

Fig.~\ref{fig:case2a} and Fig.~\ref{fig:case2b} show that the cases of emotional-shift on the IEMOCAP dataset. In Fig.~\ref{fig:case2a}, when a speaker's emotion is \textit{Neutral} for several preceding consecutive utterances, most of the models (e.g., MMGCN) tend to identify the speaker's next utterance as \textit{Neutral}. In Fig.~\ref{fig:case2b}, when a speaker's emotion was \textit{Neutral} for several consecutive utterances, the majority of models trend towards recognizing the next utterance spoken by another speaker as \textit{Neutral}. Unlike approaches such as MMGCN, our proposed GraphCFC can accurately identify the emotion of utterance as \textit{Excited} in the above two cases.

\section{Conclusion}\label{sec:conclusion}
In this paper, we propose a directed Graph based Cross-modal Feature Complementation (GraphCFC) method for reducing the multimodal heterogeneity gap and compensating the inadequacies of earlier SOTA methods such as MMGCN. Concretely, we model the multimodal dialogue as a directed graph with variable context and extract distinct types of edges from the graph for graph attention learning, thus ensuring that GNNs can select accurately critical intra-modal contextual and inter-modal interactive information; meanwhile, we also address the heterogeneity gap using multiple subspace mapping functions and PairCC strategy. In addition, we design a graph-based network structure named GAT-MLP, which provides a feasible solution for multimodal interaction. Experimental results on the IEMOCAP and MELD datasets show that our proposed GraphCFC outperforms other SOTA methods and is capable of effectively modeling long-distance intra-modal contextual information and inter-modal interactive information. 

As we can see, some challenges of multimodal machine learning remain. In future work, we hope to further explore the methodologies of multimodal fusion and evaluate the validity of the GAT-MLP layer on other multimodal tasks. Another work in the future is how to alleviate the problem of class imbalance and background noise on the MELD dataset in the conversational emotion recognition task.

\bibliographystyle{IEEEtran}
\balance
\bibliography{graphcfc}

\end{document}